\documentclass[11pt]{article}

\usepackage{preprint}
\usepackage[numbers,comma,sort&compress]{natbib} 
\usepackage[colorlinks=true]{hyperref}

\usepackage{tabularx}
\usepackage{booktabs}
\usepackage{makecell}
\usepackage{subcaption}
\usepackage[inline]{enumitem}
\usepackage{placeins}
\usepackage{caption}
\usepackage{svg}
\usepackage{float}
\usepackage{cleveref}
\usepackage{rotating}
\usepackage{array}
\usepackage{pdflscape}
\usepackage[table,dvipsnames]{xcolor}
\usepackage{pifont}
\usepackage{threeparttable}

\Crefname{extendedtable}{Extended Data Table}{Extended Data Tables}

\usepackage{xcolor}
\definecolor{fig1_development}{HTML}{AEE1FC}
\definecolor{fig1_test}{HTML}{FFD5A5}
\definecolor{fig4_main}{HTML}{AEE1FC}
\definecolor{fig4_aug}{HTML}{FFD5A5}
\definecolor{fig5_cheat}{HTML}{FFBFBF}
\definecolor{fig6_nih}{HTML}{FFEDAB}
\definecolor{fig6_padchest}{HTML}{B25A9F}

\title{CXR-LT 2026 Challenge: Multi-Center Long-Tailed and Zero Shot Chest X-ray Classification}       

\author[1,$\dagger$]{Hexin Dong}
\author[1,$\dagger$]{Yi Lin}
\author[2]{Pengyu Zhou}
\author[3]{Fengnian Zhao}
\author[4]{Alan Clint Legasto}

\author[5]{Juno Cho}
\author[6]{Dohui Kim}
\author[7]{Justin Namuk Kim}
\author[5]{Mingeon Kim}
\author[8]{Sunwoo Kwak}
\author[9]{Gabriel Moyà-Alcover}
\author[10,11]{Ky Trung Nguyen}
\author[12]{Thanh-Huy Nguyen}
\author[11,13,14]{Ha-Hieu Pham}
\author[14,15]{Huy-Hieu Pham}
\author[10]{Huy Le Pham}
\author[16]{Nikhileswara Rao Sulake}
\author[9]{Aina Tur-Serrano}
\author[17]{Ruichi Zhang}
\author[17]{Ang Zu}

\author[18]{Adam E. Flanders}
\author[19]{Zhiyong Lu}
\author[20]{Ronald M. Summers}

\author[21]{Mingquan Lin}
\author[22]{Hao Chen}
\author[23]{Yuzhe Yang}
\author[4]{George Shih}
\author[1,4,*]{Yifan Peng}

\affil[1]{Department of Population Health Sciences, Weill Cornell Medicine, New York, USA}
\affil[2]{Department of Radiology, Fuwai Hospital, National Center for Cardiovascular Diseases, Chinese Academy of Medical Sciences and Peking Union Medical College, Beijing, China}
\affil[3]{
Department of Radiology, West China School of Medicine, Sichuan University, Sichuan University Affiliated Chengdu Second People's Hospital, Chengdu Second People's Hospital, Chengdu, China}
\affil[4]{Department of Radiology, Weill Cornell Medicine, New York, USA}

\affil[5]{School of Electrical Engineering, Korea Advanced Institute of Science and Technology, Daejeon, South Korea}
\affil[6]{Gwangju Institute of Science and Technology, Gwangju, South Korea}    
\affil[7]{Department of Biomedical Engineering, Case Western Reserve University, Ohio, USA}    
\affil[8]{School of Electrical and Computer Engineering, Cornell Tech, New York, USA}  
\affil[9]{Department of Mathematics and Computer Science, Universitat de les Illes Balears, Palma de Mallorca, Spain}
\affil[10]{School of Computer Science and Engineering, VNU-HCM International University, Ho Chi Minh City, Vietnam}  
\affil[11]{Vietnam National University, Ho Chi Minh City, Vietnam}  
\affil[12]{School of Computer Science, Carnegie Mellon University, Pittsburgh, USA}  
\affil[13]{VNU-HCM University of Science, Ho Chi Minh City, Vietnam}  
\affil[14]{VinUni-Illinois Smart Health Center, VinUniversity, Hanoi, Vietnam}  
\affil[15]{College of Engineering \& Computer Science, VinUniversity, Hanoi, Vietnam}       
\affil[16]{Department of Computer Science and Engineering, Rajiv Gandhi University of Knowledge Technologies, Nuzvid, India} 
\affil[17]{School of Informatics, Xiamen University, Xiamen, China}     
\affil[18]{Department of Radiology, Thomas Jefferson University, Pittsburgh, USA}        
\affil[19]{National Library of Medicine, National Institutes of Health, Bethesda, USA}      
\affil[20]{Clinical Center, National Institutes of Health, Bethesda, USA}     
\affil[21]{Department of Surgery, University of Minnesota Twin Cities, Minneapolis, USA} 
\affil[22]{Department of Computer Science and Engineering, The Hong Kong University of Science and Technology, Hong Kong, China} 
\affil[23]{Computational Medicine and Computer Science, University of California, Los Angeles, Los Angeles, USA}      

\affil[$\dagger$]{These authors contributed equally to this work.}
\affil[*]{Corresponding author. Email: \url{yip4002@med.cornell.edu}}

\begin{document}

\maketitle

\begin{abstract}
Chest X-ray (CXR) interpretation is hindered by the long-tailed distribution of pathologies and the open-world nature of clinical environments. Existing benchmarks often rely on closed-set classes from a single institution, failing to capture the prevalence of rare diseases or the appearance of novel findings. To address this, we present the CXR-LT challenge.
The first event, CXR-LT 2023, established a large-scale benchmark for long-tailed multi-label CXR classification and identified key challenges in rare disease recognition. CXR-LT 2024 further expanded the label space and introduced a zero-shot task to study generalization to unseen findings. Building on the success of CXR-LT 2023 and 2024, this third iteration of the benchmark introduces a multi-center dataset comprising over 145,000 images from PadChest and NIH Chest X-ray datasets. Additionally, all development and test sets in CXR-LT 2026 are annotated by radiologists, providing a more reliable and clinically grounded evaluation than report-derived labels. The challenge defines two core tasks this year: (1) Robust Multi-Label Classification on 30 known classes and (2) Open-World Generalization to 6 unseen (out-of-distribution) rare disease classes. This paper summarizes the overview of the CXR-LT 2026 challenge. We describe the data collection and annotation procedures, analyze solution strategies adopted by participating teams, and evaluate head-versus-tail performance, calibration, and cross-center generalization gaps. Our results show that vision-language foundation models improve both in-distribution and zero-shot performance, but detecting rare findings under multi-center shift remains challenging. Our study provides a foundation for developing and evaluating AI systems in realistic long-tailed and open-world clinical conditions.
\end{abstract}


\section{Introduction}

Chest X-ray (CXR) is one of the most widely used imaging modalities in clinical practice. However, the distribution of abnormal findings in CXR datasets is highly imbalanced, with a small number of common abnormalities accounting for most observations, whereas many clinically important abnormalities occur rarely~\citep{zhou2021review}. Such an imbalance poses a fundamental challenge for standard deep learning methods, which tend to bias predictions toward frequent classes while underperforming on rare but critical ``tail" classes~\citep{holste2022long}. 

Although numerous methods have been proposed to address class imbalance, most existing studies focus on algorithmic improvements under fixed datasets~\citep{zhang2021mbnm,zhang2023deep,yang2022proco,ju2021relational}. In contrast, comparatively little attention has been paid to benchmark designs that capture the combined challenges of long-tailed distributions, open-world conditions, and real-world data variability. In practice, clinical models must not only recognize common abnormalities but also generalize to infrequent and previously unseen abnormalities, often across institutions and imaging settings~\citep{oakden2020hidden}.

To advance research in this direction, we established the CXR-LT challenge series. The first edition (CXR-LT 2023~\citep{holste2023cxr,holste2024towards}) introduced a benchmark for long-tailed multi-label classification. It provided a large-scale training dataset derived from radiology reports and evaluated model performance across a diverse set of abnormal findings with highly imbalanced prevalence. The primary goal was to identify the limitations of existing deep learning models, particularly their difficulty in accurately recognizing infrequent diseases and handling multi-label predictions. The second event (CXR-LT 2024~\citep{lin2025cxr}) expanded the benchmark in two major directions. First, it significantly increased both the dataset scale and label diversity, comprising over 377,000 CXRs annotated with 45 disease categories, including 19 newly introduced rare findings. Second, it introduced a zero-shot learning task to evaluate the model's ability to recognize previously unseen abnormalities without direct supervision. This addition explicitly studied open-world settings, where models must handle diseases not observed during training.

Building on these efforts, \textbf{CXR-LT 2026} further advances the benchmark to better reflect real-world clinical conditions. First, we introduce a multi-center dataset by combining PadChest~\citep{bustos2020padchest} and NIH Chest X-ray data~\citep{summers2019nih}, enabling cross-center evaluation while addressing recent access restrictions of prior datasets~\citep{PhysioNet-mimic-cxr-2.0.0}. Second, we improve label quality by using radiologist-annotated development and test sets, which provide more reliable ground truth than report-derived labels in previous challenges. Third, we define two complementary tasks that jointly evaluate long-tailed classification of known findings and open-world generalization to unseen findings.
\begin{figure}[t!]
    \centering
    \begin{subfigure}[t]{0.48\textwidth}
        \centering
        \includegraphics[width=\linewidth]{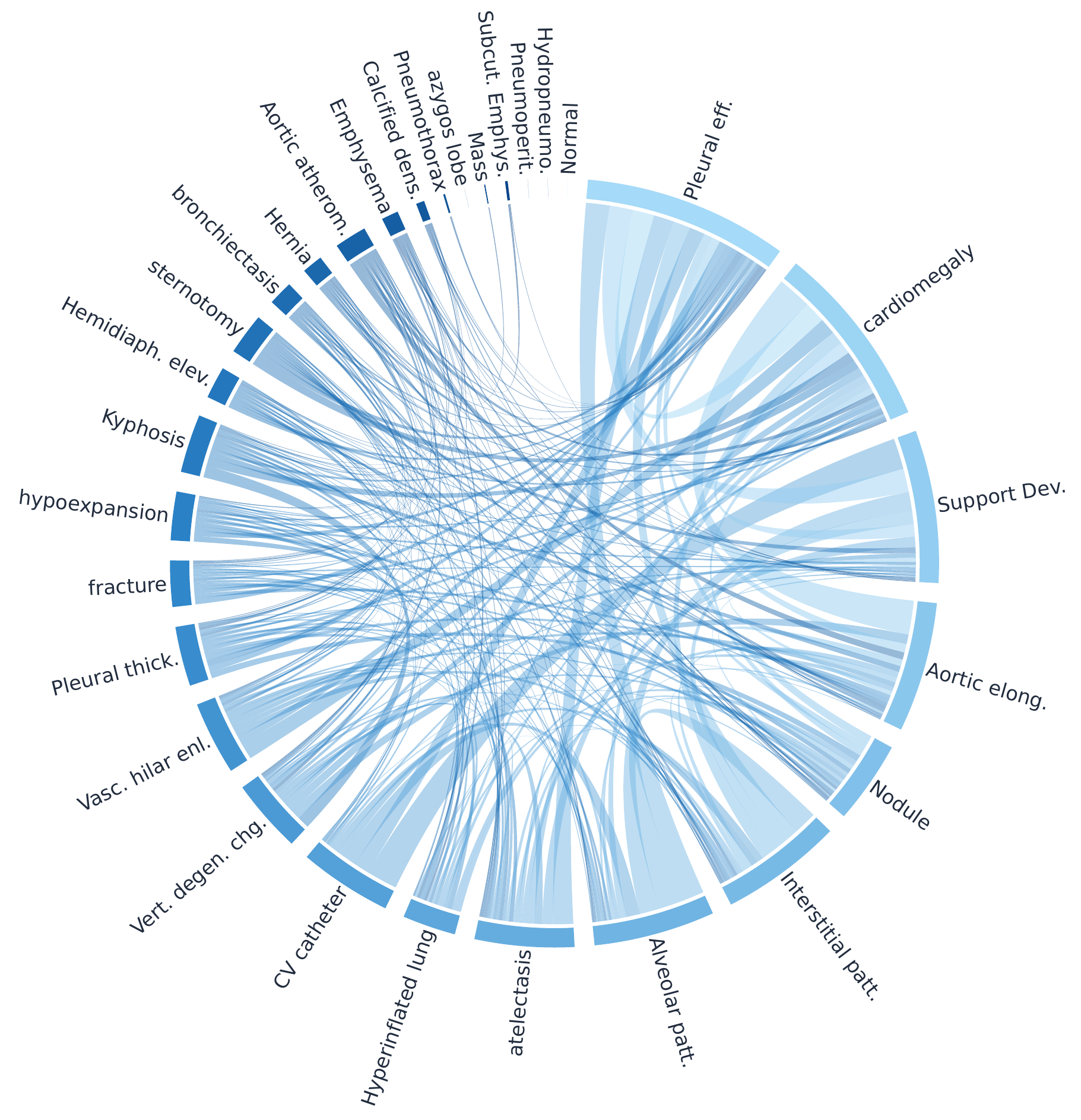}
        \caption{}
    \end{subfigure}

    \begin{subfigure}[t]{\textwidth}
        \centering
        \includegraphics[width=\linewidth]{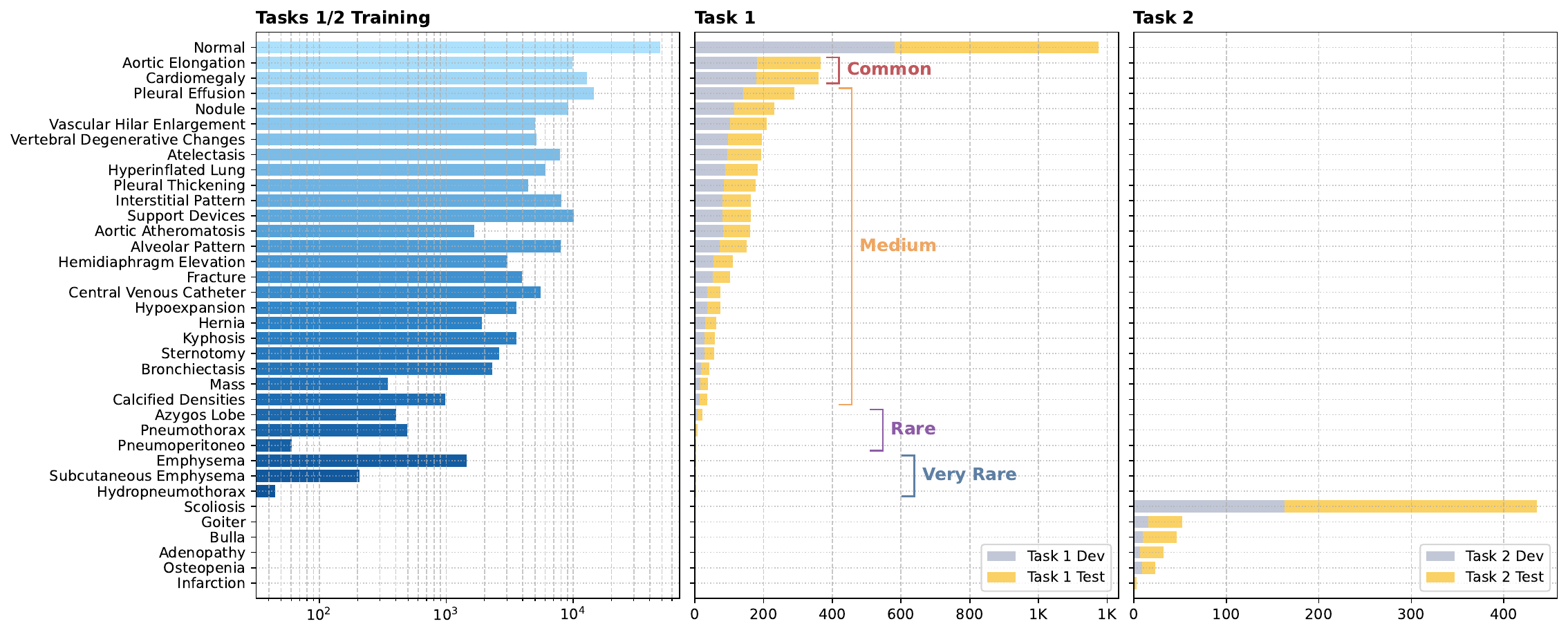}
        \caption{}
    \end{subfigure}
    \caption{The dataset of CXR-LT 2026. (a) The co-occurrence of labels in the training set. (b) The label distributions in the training set, the development and test sets of Task 1, and the development and test sets of Task 2.
    }
    \label{fig:train_data}
\end{figure}

This paper presents a comprehensive overview of the CXR-LT 2026 challenge (Figure~\ref{fig:train_data}). We detail the dataset construction and label collection, summarize methods developed by participating teams, and provide in-depth performance analyses, including head-to-tail behavior, robustness under input perturbations, and cross-center generalization. The results highlight persistent challenges in long-tailed and open-world CXR analysis and offer insights for developing more robust and clinically applicable AI systems.

\section{Challenge description}

\subsection{Task description}\label{task_description}

\subsubsection{Task 1: Long-Tailed Multi-Label Classification}

The first task evaluates models' ability to recognize common CXR findings under a long-tailed class distribution. In this task, participants are required to predict whether 30 ``seen'' disease categories are present. The label set was constructed based on two criteria. First, we retained 20 classes from previous CXR-LT editions to ensure longitudinal consistency. Second, we added 10 additional findings from the PadChest dataset to better capture clinically common abnormalities. The final label set, therefore, reflects a broad and realistic spectrum of thoracic findings:
\begin{enumerate*}[label=(\arabic*)]
\item Alveolar Pattern,
\item Aortic Atheromatosis,
\item Aortic Elongation,
\item Atelectasis,
\item Azygos Lobe,
\item Bronchiectasis, 
\item Cardiomegaly,
\item Calcified Densities,
\item Central Venous Catheter, 
\item Emphysema,
\item Fracture,
\item Hemidiaphragm Elevation, 
\item Hernia,
\item Hydropneumothorax,
\item Hyperinflated Lung, 
\item Hypoexpansion, 
\item Interstitial Pattern, 
\item Kyphosis,
\item Mass,
\item Nodule,
\item Normal,
\item Pleural Effusion,
\item Pleural Thickening,
\item Pneumothorax,
\item Pneumoperitoneum,
\item Sternotomy,
\item Subcutaneous Emphysema,
\item Support Devices,
\item Vascular Hilar Enlargement, 
\item Vertebral Degenerative Changes.
\end{enumerate*}

These categories cover structural abnormalities, pulmonary diseases, and medical devices, spanning both frequent and infrequent findings.
Models were trained using the large-scale PadChest training dataset with report-derived labels. Evaluation was conducted on the radiologist-annotated PadChest-GR dataset~\citep{de2025padchest} to ensure a more reliable assessment compared to prior editions.

\subsubsection{Task 2: Open-World Generalization}

The second task evaluates models' ability to recognize previously unseen disease categories without explicit supervision.
The label set was derived from PadChest-GR and built upon the out-of-distribution (OOD) taxonomy introduced in Task~3 of the CXR-LT 2024. To maintain comparability, we retained the same five OOD categories and introduced one additional finding, Goiter. The final set includes:
\begin{enumerate*}[label=(\arabic*)]
\item Adenopathy,
\item Bulla, 
\item Goiter,
\item Infarction, 
\item Osteopenia, 
\item Scoliosis.
\end{enumerate*}

During training, models used the same dataset as in Task 1 but did not receive labels for these six findings.
The goal is to assess whether models can leverage learned visual representations to generalize to new disease categories. This setup reflects a realistic clinical deployment in which models may encounter abnormalities that were not explicitly annotated during training.

Evaluation was conducted on both an \textbf{internal} test set from PadChest-GR and an \textbf{external} test set from the NIH ChestX-ray dataset. The inclusion of external data enables assessment under cross-center distribution shifts, further testing model robustness. Compared with Task 1, this task is substantially more challenging, as it requires both generalization to unseen classes and stability across different data sources.

\subsection{Dataset description}

\begin{table}[t]
\centering
\footnotesize
\caption{Overview of the CXR-LT 2026 dataset.}
\label{tab:cxrlt2026_dataset}
\begin{tabular}{cllrl}
\toprule
\textbf{Tasks}& \textbf{Dataset}& \textbf{Source} & \textbf{\# CXRs} & \textbf{Curation}  \\
\midrule
1 &  Training & PadChest & 142,928 & From reports\\
  &  Development & PadChest-GR & 1,620 & By radiologists\\
  &  Test & PadChest-GR & 1,400 & By radiologists\\
\midrule
2 &  Training & PadChest & 142,928 & From reports\\
  &  Development & PadChest-GR & 200 & By radiologists\\
  &  Internal Test & PadChest-GR & 305 & By radiologists\\
  &  External Test & NIH Chest X-ray & 79 & By radiologists\\
\bottomrule
\end{tabular}
\end{table}

The CXR-LT 2026 dataset was constructed from two publicly available sources: PadChest~\citep{bustos2020padchest} and NIH ChestX-ray~\citep{summers2019nih}. It is designed to support both large-scale training under weak supervision and rigorous evaluation under clinically realistic conditions, including cross-center variability (\Cref{tab:cxrlt2026_dataset}).

\paragraph{Training data (PadChest).}

PadChest serves as the primary data source for model training. We use 142,928 CXR images, each paired with a Spanish radiology report. Disease labels were automatically extracted from these reports, resulting in a large-scale dataset with weak supervision. This setup reflects real-world conditions, where annotations are available at scale but may be noisy. Because PadChest-GR is derived from PadChest, we performed both patient-level and study-level de-duplication to prevent leakage between the weakly labeled training split and the radiologist-annotated evaluation sets.

\paragraph{Internal test (PadChest-GR).}

To provide a reliable evaluation, we use PadChest-GR~\citep{de2025padchest}, a radiologist-annotated subset of PadChest. In this dataset, multiple radiologists provided image-level annotations for each case. From this subset, we selected 3,020 frontal CXRs that contain at least one target disease finding. These images were split into development and test sets using stratified sampling to preserve the long-tailed class distribution.

\paragraph{External test (NIH ChestX-ray).}

To evaluate generalization, we constructed an additional test set from the NIH ChestX-ray dataset. This dataset differs from PadChest in acquisition protocols, patient populations, and labeling characteristics, introducing a natural distribution shift. It therefore enables evaluation of model robustness across institutions.

\subsection{Ethics, de-identification, and data access.}
\textbf{Ethics Statement}. This study exclusively uses publicly available, de-identified datasets, including PadChest, PadChest-GR, and the NIH ChestX-ray dataset. Researchers accessing the PadChest and PadChest-GR datasets are required to comply with the corresponding Data Use Agreement (DUA) and all applicable terms governing their use. No new patient data were collected, and no identifiable protected health information was accessed. Accordingly, this study does not constitute human subjects research requiring Institutional Review Board (IRB) review or approval.

All data elements released to challenge participants were de-identified. Participants received de-identified CXR images, challenge-specific image identifiers, split information, image-level training labels, and prediction-template/evaluation files. The released training labels were structured binary labels and did not contain free-text clinical information or protected health information. Patient identifiers and hidden evaluation labels were not released to participants. Radiologist annotations for the development and test sets were used internally to construct benchmarks and centralized evaluation. Radiologist annotations and hidden evaluation labels were retained internally for benchmark construction and centralized evaluation and were not released during the challenge. Participants accessed the data through registered challenge accounts and were required to follow the applicable data-use restrictions, including research-only use, no redistribution, and no attempts to re-identify individual patients.

\subsection{Label collection}

Label collection plays a central role in the CXR-LT 2026 benchmark. The dataset follows a two-stage annotation strategy that combines large-scale weak supervision for model development with expert-level annotations for evaluation.

For the \textbf{training set}, CXR-LT 2026 used the structured English finding labels released with PadChest, which were derived from the original Spanish radiology reports by the PadChest dataset creators. We did not re-run free-text extraction from the Spanish reports for this challenge. Instead, the released PadChest structured labels were used as the starting point for constructing the CXR-LT 2026 weak training labels. To align the PadChest label space with the CXR-LT 2026 taxonomy, one radiologist (P.Z.) reviewed the PadChest and PadChest-GR finding categories and curated the mapping from PadChest source labels to the 30 Task~1 categories described in \Cref{task_description}. Synonymous terms, spelling variants, and clinically overlapping source labels were collapsed into the corresponding CXR-LT category when appropriate. Source labels that lacked a clear equivalent among the 30 Task~1 categories were left unmapped.

For each training image, we generated a 30-dimensional binary multi-label vector. A category was assigned a positive label if at least one structured PadChest label for that image mapped to the corresponding CXR-LT category. If multiple PadChest source labels mapped to the same category, they were merged into a single positive label. Categories without a mapped positive PadChest label were assigned negative values. Images containing abnormal findings outside the 30 Task~1 categories but with no mapped target finding were therefore represented by all-zero vectors. These all-zero vectors indicate that no positive label was present among the 30 challenge categories and should not be interpreted as confirmed normal studies. The released training labels were binary and did not encode a separate uncertainty state; thus, uncertainty and ambiguity inherited from the original report-derived PadChest labels remain potential sources of weak-label noise.

For the \textbf{development and internal test sets}, we derived image-level labels from the radiologist-reviewed PadChest-GR annotations~\citep{de2025padchest}. PadChest-GR is a manually curated subset of PadChest in which positive findings were extracted from the original Spanish radiology reports, translated into English, linked to standardized finding labels, and manually reviewed by radiologists. The PadChest-GR annotation involved nine senior and five junior radiologists from University Hospital Sant Joan d’Alacant, Spain. Annotation was performed in two stages: study-level quality control followed by finding-level annotation. In both stages, each study or finding was independently analyzed by two professionals. During quality control, radiologists reviewed the image, report, the extracted list of findings, and, when applicable, the available prior image. Cases without consensus during quality control were re-evaluated by three senior radiologists and required approval by at least two of them. For finding-level annotation, each positive finding sentence was reviewed independently by two radiologists; localizable findings were annotated with bounding boxes, while diffuse or global findings were retained as findings but were not required to have bounding boxes when precise localization was not appropriate.

To adapt PadChest-GR to the CXR-LT 2026 challenge taxonomy, one radiologist (P.Z.) reviewed the PadChest-GR finding categories and curated the mapping from the original PadChest-GR labels to the challenge label space, rather than generating new image-level annotations from scratch. As described in \Cref{task_description}, the Task 1 label set consisted of 30 “seen” categories, including 20 categories retained from previous CXR-LT editions to maintain longitudinal consistency and 10 additional findings selected from PadChest to better capture clinically relevant abnormalities. For Task 2, we retained the five out-of-distribution categories introduced in CXR-LT 2024 and added one new category, Goiter, bringing the total to six unseen categories. We then selected PadChest-GR cases that contained at least one of these 36 target categories. The selected cases were split into development and test sets using stratified sampling to preserve the long-tailed class distribution.

For the \textbf{external test set} constructed from the NIH ChestX-ray dataset, we used a two-stage radiologist adjudication protocol implemented in \url{https://md.ai/}. Two radiologists (P.Z. [5 years of experience] and F.Z. [5 years of experience]) independently reviewed all 159 candidate images in the annotation interface. The readers were provided with the six Task 2 label definitions and were instructed to assign image-level binary labels indicating the presence or absence of the target finding(s). Cases for which both readers confirmed at least one Task 2 target finding were retained directly for the external test set (n = 30). Cases with discordant initial interpretations between P.Z. and F.Z. (one positive and one negative; n = 53) were further reviewed by a third radiologist, A.C.L. [21 years of experience], using the same MD.ai annotation interface and label definitions. Among these discordant cases, 49 were confirmed and retained. The final NIH external test set therefore consisted of 79 CXR cases, including 30 two-reader concordant positives and 49 adjudicated positives. The remaining candidates were not included in the final external test set. The annotation and agreement/adjudication flow is summarized in  \Cref{tab:annotation_protocol}. Because this external test set was intentionally enriched for rare Task 2 findings and contains a small number of cases, we report agreement/adjudication counts rather than relying only on chance-corrected inter-rater agreement statistics, which can be unstable under severe class imbalance.

\subsection{Metric}

CXR-LT 2026 focuses on multi-label classification under severe class imbalance. In this setting, standard evaluation metrics such as AUROC (Area Under the Receiver Operating Characteristic Curve) can be misleading because performance is heavily influenced by the large number of negative samples in long-tailed datasets. To address this issue, we use mean Average Precision (mAP) as the primary evaluation metric \citep{everingham2010pascal}. Specifically, we computed the average precision independently for each class and reported the macro-average across all classes as the final score. This metric evaluates the ranking quality of predicted probabilities across all decision thresholds and provides a more reliable measure of performance in long-tailed multi-label settings. 

In addition to mAP, we report several auxiliary metrics to provide complementary insights into model behavior:
\begin{enumerate*}[label=(\arabic*)]
\item mean AUROC (mAUROC) to measure ranking performance,
\item mean F1 score (mF1), mean precision, and mean recall, computed using a fixed decision threshold, and
\item Expected Calibration Error (ECE) to assess the reliability of predicted probabilities.
\end{enumerate*}
These auxiliary metrics are presented on the leaderboard, but only mAP is used for final ranking.

\subsection{Challenge timeline}

The CXR-LT 2026 challenge was organized in a staged timeline to ensure fair evaluation while supporting iterative model development. 

\paragraph{Training phase (Nov 24, 2025).}

The training set was released, allowing participants to explore the dataset, design models, and develop baseline approaches. Teams were required to register for the challenge to obtain access to the data.

\paragraph{Development phase (Dec 1, 2025 - Jan 19, 2026).}

The development set was made available for iterative evaluation. Participants could submit predictions and access immediate feedback via the online CondaBench platform\footnote{
Task 1: \url{https://www.codabench.org/competitions/11470/}; 
Task 2: \url{https://www.codabench.org/competitions/11471/}.
}. This phase enabled performance analysis, error diagnosis, and model refinement.

\paragraph{Final evaluation phase (Jan 20, 2026 – Feb 4, 2026).}

The final test set was released under `blind test' rules. Specifically, participants were allowed to submit up to 5 valid predictions, without access to ground-truth labels or intermediate results. All submissions were evaluated centrally. 

\subsection{Evaluation protocol}\label{sec:Evaluation}

Final rankings were determined based on the primary metric (mAP). Secondary metrics (i.e., mAUC, mF1, ECE, mean precision, and mean recall) were reported for additional insights and used only to break ties when teams have the same mAP scores.

To ensure fairness while allowing participants to use contemporary modeling strategies, the challenge permitted training on external data and the use of publicly available pretrained models, provided that neither these data nor models were trained on images from the CXR-LT 2026 development or test sets. For example, models pretrained on general-domain datasets such as ImageNet, or on external CXR datasets that do not overlap with the CXR-LT evaluation images, were allowed. In contrast, participants were instructed to avoid pretrained models based on PadChest when the pretraining corpus could include images from the competition development or test sets, because the internal evaluation sets were derived from PadChest-GR. Participants were also not allowed to use hidden test labels, manually relabel evaluation images, or use external annotations overlapping with the CXR-LT evaluation sets.

For Task 2, a stricter rule was applied to preserve the open-world setting. Participants were not allowed to use any images containing the six OOD target labels for supervised training, nor to use pretrained models obtained using these images or similar OOD-label supervision. These six OOD-related labels were therefore required to remain unseen during supervised training. Participants could use disease names, textual prompts, and general or medical vision--language representations for zero-shot or prompt-based inference, but they could not train or fine-tune models using supervised image-level examples of the six Task 2 categories.

In addition to the official evaluation, we conducted an auxiliary
robustness analysis to assess model sensitivity to controlled input
perturbations. Each image in the evaluation set was paired with a
single perturbed counterpart generated once using a prespecified
stochastic pipeline. The pipeline consisted of three operations:
(1) affine scaling, with the scale factor randomly selected from
$\{0.9, 1.1\}$; (2) affine translation, with the displacement randomly
selected as either $-10\%$ or $+10\%$ of the image dimension along
each spatial axis; and (3) additive Gaussian noise, applied with a
probability of 0.5, with the standard deviation sampled from the
range $[0.1, 0.2]$. Horizontal flipping was not included in the final
perturbation pipeline.

One perturbed realization was generated for each image, and no
repeated perturbation trials were performed. The resulting perturbed
evaluation set was fixed and used for all submitted models. Each model
was evaluated on the original and perturbed sets separately using the
same preprocessing, metrics, and evaluation code. Predictions from
the original and perturbed images were not averaged or otherwise
combined; therefore, this analysis should not be interpreted as
multi-view test-time augmentation. The official challenge ranking was
based only on the original test set, and the perturbation analysis was
reported as an auxiliary robustness evaluation. Confidence intervals
were estimated using the same 1,000 case-level bootstrap procedure
described in Section~\ref{sec:primary_results}.

\section{Results}
\subsection{Team participation across stages}

A total of 72 teams initially registered and obtained access to the training data. During the development phase, 23 teams submitted predictions to the public leaderboard, resulting in 2,475 submissions in total. This high submission volume reflects sustained engagement and iterative experimentation throughout model development. 

\begin{figure}
\centering
\includegraphics[width=0.8\textwidth]{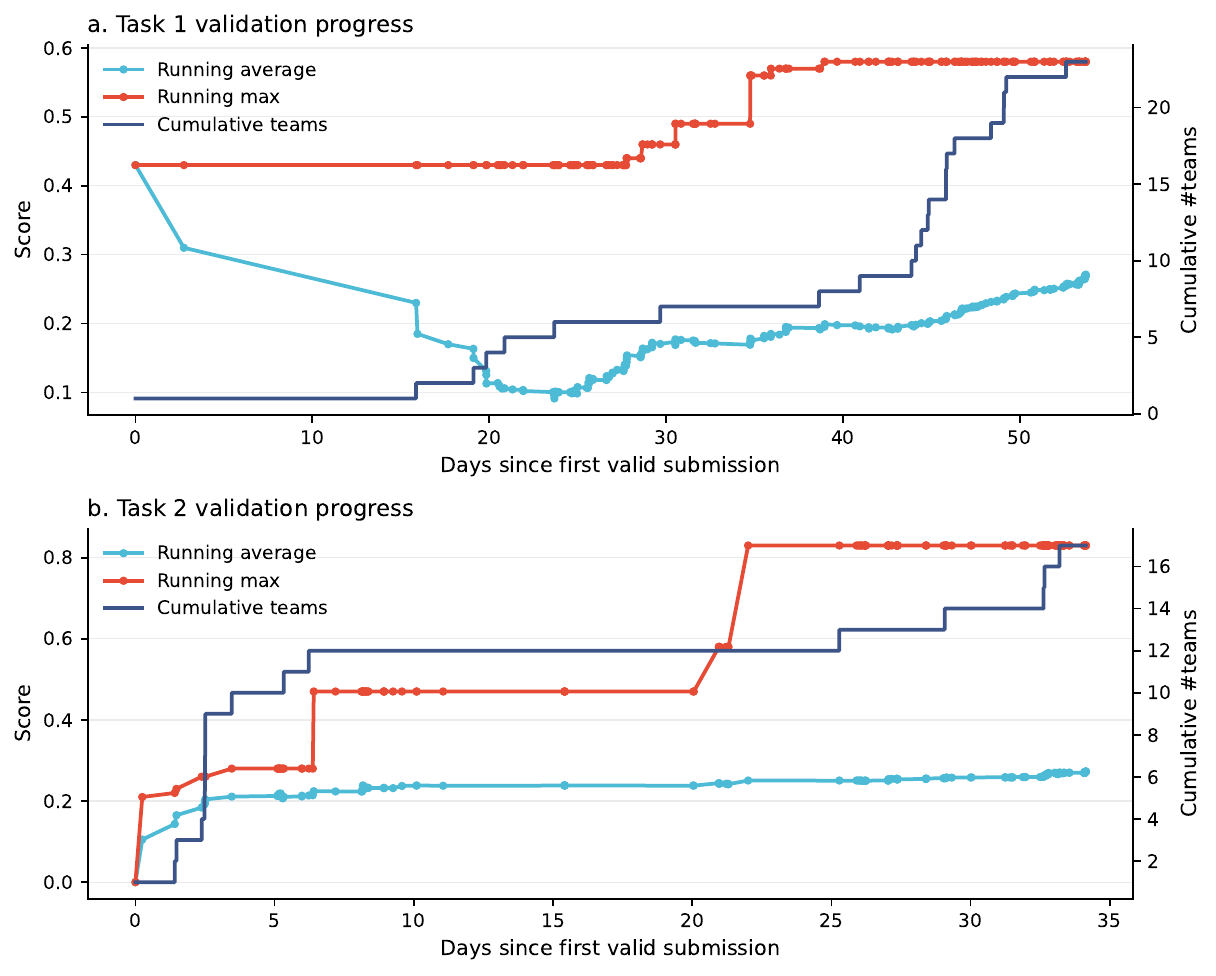}
\caption{Validation progress over time for Task 1 and Task 2.}
\label{fig:task_progress}
\end{figure}

As shown in \Cref{fig:task_progress}, model performance improved steadily over time. The running maximum score exhibits several distinct jumps, indicating moments when new methodological advances or refinements led to notable performance gains. In contrast, the running average score increases more gradually, suggesting consistent incremental improvements across participating teams. In addition, the number of active teams increased during the later stages of the development phase, reflecting growing engagement as participants refined their approaches. Similar trends are observed in both tasks, although performance improvements in Task 2 appear more abrupt, highlighting the greater difficulty of the open-world setting.

At the final evaluation stage, 21 teams submitted complete test-set predictions for official ranking. Following the release of final results, all eligible teams were required to submit technical reports describing their methodology. Only teams that completed this final submission were included in the official challenge summary.

\subsection{Participating methods}

\Cref{tab:teams_task1} summarizes the top-performing methods for both Task 1 and Task 2. For each team, we report key components of their approach, including image resolution, model backbone, training strategy, pretraining dataset, and inference strategy. 
A clear methodological distinction emerges between the two tasks. For Task 1, most methods rely on convolutional neural networks or transformers combined with techniques such as loss reweighting or sampling strategies to mitigate the effects of long-tailed class distributions.
In contrast, Task 2 is dominated by vision-language models, which leverage joint image–text representations for open-world generalization. These approaches typically employ prompt-based inference or metric-based matching strategies.

\begin{table}[t]
\caption{
Summary of the top-performing methods for Task 1 and Task 2.
}
\label{tab:teams_task1}
\footnotesize
\begin{tabularx}{\linewidth}{l>{\raggedright\arraybackslash}p{2.5cm}l>{\raggedright\arraybackslash}p{2cm}>{\raggedright\arraybackslash}X>{\raggedright\arraybackslash}p{2cm}>{\raggedright\arraybackslash}p{2cm}}
\toprule
Team & Institution & \makecell[cl]{Image\\Resolution} & Backbone & Strategy &  Pretraining & \makecell[cl]{Inference\\Strategy}\\
\midrule
\multicolumn{7}{l}{\cellcolor{GreenYellow}{Task 1}}\\
A & VinUni-Illinois Smart Health Center  & 512 & ConvNextv2-Base &  Distribution-Balanced loss &  MIMIC-CXR & TTA, Model ensemble \\
\rowcolor{gray!20}
B & Korea Advanced Institute of Science and Technology & 512 & ConvNeXtV2, SwinV2-T & - & PCAM pretraining&TTA\\
C &Vietnam National University & 384 & ConvNeXt & Asymmetric Loss, Auxiliary Loss & ImageNet &TTA  \\
\rowcolor{gray!20}
D & Cornell University &  768 & Swin-L & Asymmetric loss & Ark+ Checkpoint & TTA \\
E & Rajiv Gandhi University of Knowledge Technologies & 512 & ConvNext-Large & LDAM, class-balanced sampling & ImageNet & TTA\\
\rowcolor{gray!20}
F & Universitat de les Illes Balears& 224 & DenseNet121 & Asymmetric Loss, Multi-Supervised Contrastive Loss &CheXpert & -\\
\midrule
\multicolumn{7}{l}{\cellcolor{GreenYellow}{Task 2}}\\
A & VinUni-Illinois Smart Health Center & 224 & OpenCLIP ViT-L/14 & Prompt engineering\& prompt ensembling & MIMIC-CXR &Metric-based inference  \\
\rowcolor{gray!20}
B & Korea Advanced Institute of Science and Technology & 224 & CheXzero & Asymmetric Loss & MIMIC-CXR, CheXpert & TTA, Prompt ensembling\\
G & Xiamen University & 224 & MedKlip, Resnet50, ClinicalBERT & Class-balanced loss & MIMIC-CXR, CheXpert & Prototype Prompting  \\
\bottomrule
\end{tabularx}
\end{table}

\paragraph{Team A: CVMAIL x MIHL.} 

This team participated in both Task 1 (long-tailed multi-label classification) and Task 2 (zero-shot OOD recognition) of the CXR-LT 2026 challenge \citep{pham2026handling}. For Task 1, they develop an imbalance-aware multi-label learning framework based on ConvNeXtV2-Base \citep{woo2023convnext}. To mitigate extreme class imbalance, they adopt a Distribution-Balanced loss with effective-number reweighting and a positive logit-margin adjustment to enhance tail recognition while maintaining stability on head classes. Class-Aware Sampling further increases exposure to rare findings. During inference, they apply test-time augmentation and weighted checkpoint ensembling, followed by a lightweight normal-gating refinement to suppress spurious abnormal predictions. For Task 2, they employ WhyXrayCLIP \citep{yang2024textbook}, a CXR-specific vision-language model built on top of OpenCLIP ViT-L/14 \citep{cherti2023reproducible}. Zero-shot recognition is formulated as image–text similarity matching in a shared embedding space, where unseen disease categories are represented using textual prompts. Prompt ensembling improves robustness to wording variations, and no supervised OOD labels are used during training. This unified framework addresses supervision scarcity by combining imbalance-aware optimization for in-distribution learning with contrastive vision-language alignment for zero-shot generalization.

\paragraph{Team B: Cool Peace.} 

This team proposed two independent models for CXR-LT classification: one for Task 1 and one for Task 2 \citep{cho2026cxr}. For multi-label classification (Task 1), the team trained an EfficientNet Router to predict which projection an input image belongs to. Based on this prediction, the input image is fed into a corresponding projection-aware model, each trained only on a single projection. The final logit for each branch is obtained by ensembling the outputs of three models. Each ensemble component uses modern visual encoder architectures, such as ConvNeXt, Class Attention Image Transformer (CaiT), Shifted-Window (Swin) Transformer \citep{liu2021swin}, and their hybrids \citep{park2022multi}. The team adopted strong data augmentation, including rotation, horizontal flipping, scaling, and shifting, to enhance robustness. Test-Time Augmentation (TTA) was also applied at the inference stage. For zero-shot classification (Task 2), the team utilized CheXzero \citep{tiu2022expert} as the base vision-language model. To address the severe long-tailed distribution in the PadChest dataset, the architecture was expanded into a dual-branch structure. A classification branch was integrated alongside the original contrastive branch, computing cosine similarities between image and disease description embeddings to generate logits supervised by Asymmetric Loss (ASL)~\citep{ridnik2021asymmetric}. The optimal loss weighting between the two branches was determined via a proxy Out-Of-Distribution (OOD) validation strategy, curating three distinct experimental sets by splitting the 30 available training classes into 24 seen and 6 unseen classes. Additionally, LLM-generated prompt descriptions were used to extract context-aware text embeddings. During inference, text shuffling and TTA were applied to improve generalizability.

\paragraph{Team C: VIU.} 

This team takes part in Task 1 of the ISBI 2026 CXR-LT challenge \citep{Ky2604}. They adopt a lightweight two-phase training framework with a Mixture-of-Experts (MoE) integration strategy to enhance the model performance on the multi-label CXR-LT challenge, specifically in task 1. In the first phase, they fully train a ConvNeXt-B \citep{liu2022convnet} backbone together with a query-based decoder classification head \citep{liu2021query2label}. In the second phase, they freeze most backbone parameters and train only the last layers and the classification head. Most importantly, they further incorporate MoE into the architecture by replacing the MLP layer of the last ConvNeXt layers and the decoder head. Through this MoE mechanism, they aim to promote expert specialization across multiple clinical diseases rather than relying on a single shared representation, thereby further improving the model's generalization. Eventually, they employ an asymmetric loss~\citep{ridnik2021asymmetric} to mitigate the data imbalance, followed by an auxiliary loss to prevent expert collapse.

\paragraph{Team D: Bibimbap-Bueno.}

This team utilized the Ark+ backbone \citep{ma2025fully} for CXR-LT classification in Task 1. Their approach used the Ark+ backbone pretrained on over 700,000 chest X-ray images as the feature extractor. Specifically, the backbone weights of the Swin-Large model \citep{liu2021swin} were initialized from Ark+ pretrained weights and fine-tuned end-to-end on the challenge dataset. To address the severe class imbalance, they used asymmetric loss (ASL)~\citep{ridnik2021asymmetric}, which downweights easy negatives and focuses learning on rare positive findings in the long tail. Test-time augmentation was implemented via multi-crop evaluation (n=10) using multiple spatial crops and their horizontal flips with prediction averaging.

\paragraph{Team E: Nikhil Rao.}

This team participated in Task 1. The approach systematically evaluates imbalance-aware loss functions and modern CNN architectures on the CXR-LT 2026 benchmark. The primary strategy employed Label-Distribution-Aware Margin (LDAM) loss \citep{cao2019learning} with Deferred Re-Weighting (DRW) to improve minority-class margins. Multiple backbones were explored, including ResNet, DenseNet, EfficientFormerV2, and ConvNeXt \citep{liu2022convnet} variants. A two-stage classifier retraining (cRT) \citep{kang2019decoupling} strategy was applied to decouple representation learning from classifier optimization. Test-time augmentation (horizontal flips and small rotations) and weighted probability ensembling were further used to enhance robustness. ConvNeXt-Large achieved the strongest single-model performance, while ConvNeXt-Base with cRT+TTA improved ranking metrics. A weighted ensemble of ConvNeXt-Large, ConvNeXt-Base, and EfficientNetV2-S achieved 52\% mAP on the development set.

\paragraph{Team F: UGIVIA.} 

This team developed a two-stage framework for CXR-LT Task 1. Their approach first employed a binary DenseNet121 model to distinguish normal from pathological chest X-rays, then a multi-label DenseNet121 to predict specific diseases. Both models used CheXpert-
pretrained weights from TorchXRayVision \citep{cohen2022torchxrayvision}, with DenseBlock 3-4 finetuned. To handle long-tailed distributions, the multi-label model included a projection head optimizing a weighted combination of Asymmetric Loss ($\gamma_{neg}=6$, $\gamma_{pos}=1$) \citep{ridnik2021asymmetric} and Multi-Supervised Contrastive Loss ($\alpha$ = 1.0, $\beta$ = 1.0, temperature = 0.07) \citep{audibert2024multi}. To address class imbalance, they applied targeted augmentations (rotations and horizontal flips) for classes with fewer than 4,000 samples and multilabel-stratified splitting. During inference, the binary classifier predicted images as normal or diseased, forwarding pathological images to the multi-label model, while normal images were directly labeled normal.

\paragraph{Team G: Zuang.} 
This team proposes a framework for long-tailed multi-label chest X-ray classification and zero-shot recognition of unseen diseases and participates in Task 2 of the CXR-LT challenge. Zero-shot recognition of unseen findings in long-tailed multi-label chest X-ray classification remains a fundamental challenge. To address this, this team proposes a method that combines model fine-tuning with a class-prototype
prompting mechanism. Semantic-Aware Vision–Language Pretraining leverages a medical vision–language model to align chest X-ray images with structured disease label representations, enabling generalization to unseen categories \citep{wu2023medklip}. Domain- Adaptive Fine-Tuning adapts the model to the target CXR dataset, enhancing discriminative feature learning while preserving cross-modal semantic knowledge to better handle long-tailed distributions. Finally, Prototype-Based Zero-Shot Inference constructs robust class prototypes from high-confidence samples, providing stable semantic anchors that guide zero-shot prediction for unseen diseases \citep{pourpanah2022review}. Extensive experiments validate the effectiveness of our framework, achieving a mAP of 0.2235 and ranking third among all participants.

\subsection{Primary evaluation results}\label{sec:primary_results}

\begin{figure}
    \begin{subfigure}[t]{0.45\textwidth}
        \centering
        \includegraphics[width=.9\textwidth]{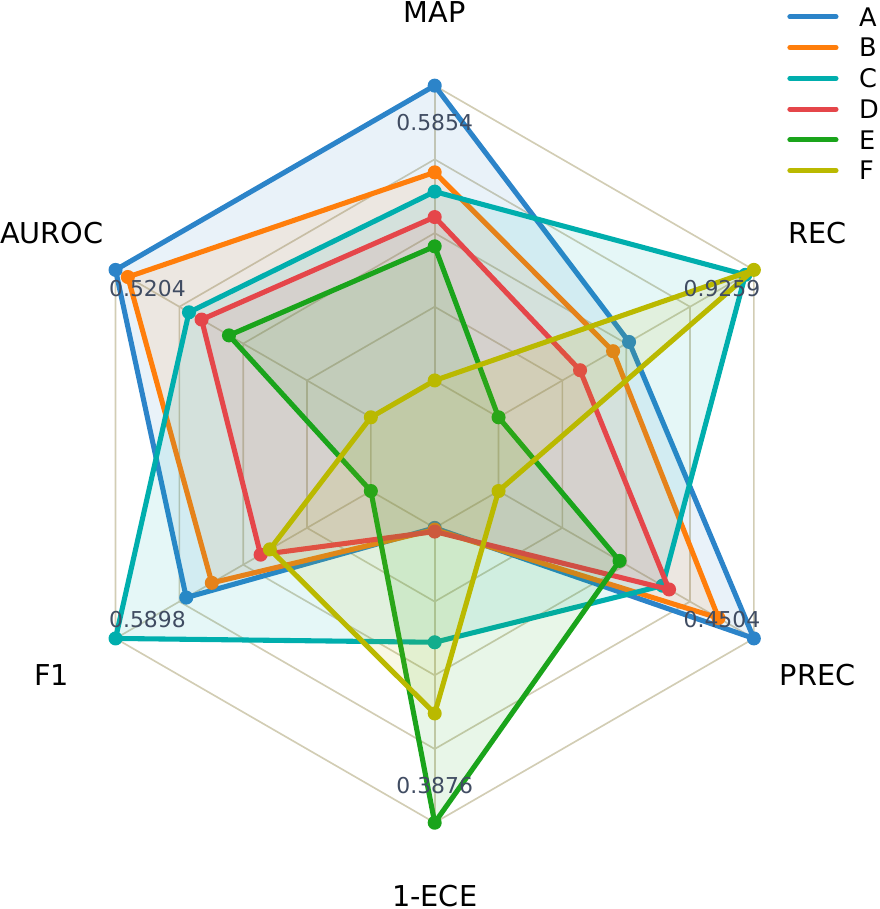}
        \caption{Task 1}
        \label{fig:radar_task1}
    \end{subfigure}
    \hfill
    \begin{subfigure}[t]{0.45\textwidth}
        \centering
        \includegraphics[width=.9\textwidth]{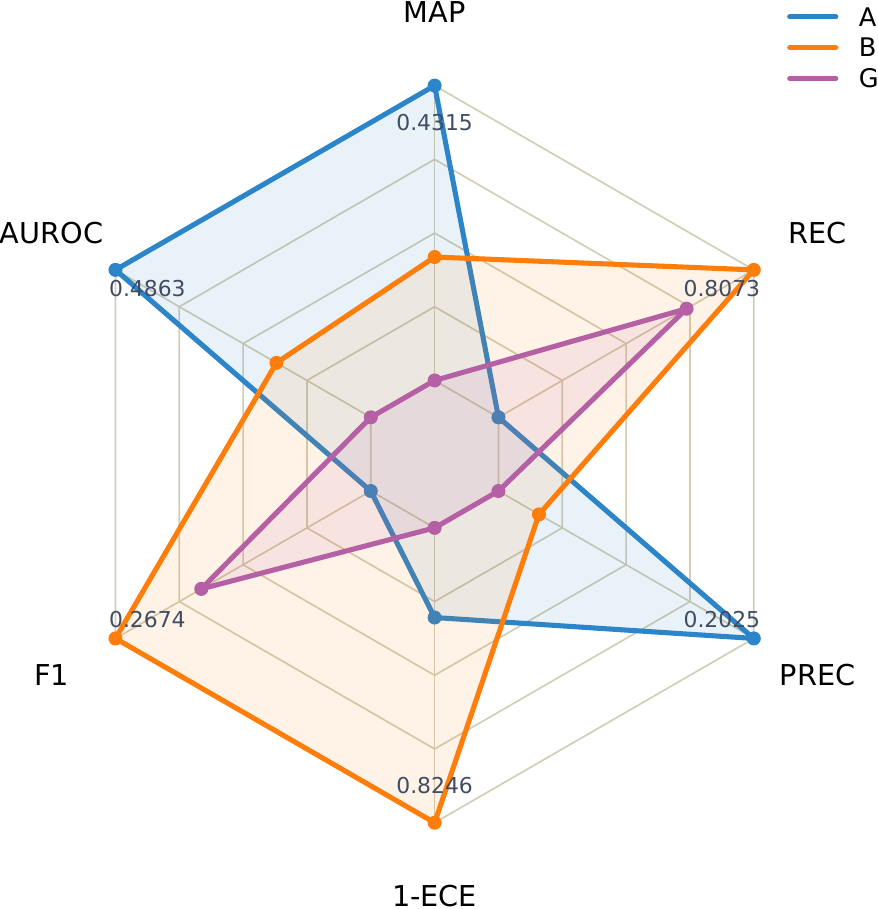}
        \caption{Task 2}
        \label{fig:radar_task2}
    \end{subfigure}
\caption{
Radar plots summarizing the main results of CXR-LT 2026 for the two challenge tasks. 
\textbf{a}, Task 1 evaluates long-tailed multi-label classification on seen classes. 
\textbf{b}, Task 2 evaluates generalization to unseen rare findings in an open-world setting. 
The plots compare the leading methods across six evaluation metrics: mAP, AUROC, F1, $1-ECE$, precision, and recall. Please note we report \emph{$1 - ECE$} as a direction-aligned calibration score, such that higher values indicate better calibration.
}
\label{fig:main_radar_results}
\end{figure}

The primary results of CXR-LT 2026 are shown in \Cref{fig:main_radar_results} and detailed in  \Cref{tab:task1_main_ci}. Model performance is primarily evaluated using mean Average Precision (mAP) for ranking, while additional metrics (AUROC, F1 score, ECE, precision, and recall) are reported to provide complementary perspectives on model behavior. Confidence intervals were estimated using 1,000 case-level bootstrap resamples, with each case corresponding to a CXR image. In each run, cases were sampled with replacement from the corresponding evaluation set, and all labels and prediction scores for each sampled case were retained together. Metrics were recomputed on each bootstrap sample, and 95\% confidence intervals were obtained from the 2.5th and 97.5th percentiles of the bootstrap distribution. No class-level stratification was applied, because labels can co-occur within the same image. Classes with no positive cases, or an undefined AUROC due to only one observed class, were excluded from the macro-average for that bootstrap sample. For cross-center analyses, resampling was performed separately within the PadChest-GR and NIH ChestX-ray subsets.

Task 1 evaluates multi-label classification on 30 known abnormal findings under a long-tailed class distribution. The top-performing method achieves an mAP of 0.5854, with a clear margin over the other submissions. AUROC scores are generally high across teams, indicating that most methods can effectively distinguish positive and negative samples at a ranking level. However, substantial differences appear in threshold-dependent metrics. In particular, some methods achieve higher F1 and recall despite having lower ranking metrics, suggesting that their predictions may operate at different confidence thresholds. This discrepancy indicates that strong ranking performance does not necessarily translate into optimal binary classification outcomes. 
Calibration performance also varies considerably across teams. Although achieving high AUROC scores, several methods exhibit relatively large ECE, indicating that their predicted probabilities are not well aligned with true outcome frequencies. This limitation is important in clinical settings, where reliable probability estimates are critical for decision support.

Task~2 evaluates model generalization to six unseen disease categories not included in the training data. Compared with Task 1, performance drops substantially across all metrics, reflecting the challenge of recognizing previously unseen abnormalities. The best-performing method achieves an mAP of 0.4315, while the remaining methods show noticeably lower performance. This gap indicates that current models still struggle to generalize beyond the label space seen during training. Consistent with Task 1, ranking-based and threshold-dependent metrics do not always align. For example, some methods achieve higher recall despite lower mAP, suggesting different trade-offs between sensitivity and ranking quality. 

\Cref{fig:main_radar_results} also reveals three consistent observations. First, large performance gaps remain between the top-performing methods and the remaining submissions, indicating substantial differences in how models handle long-tailed distributions. Second, ranking-based metrics such as mAP and AUROC do not always align with threshold-dependent metrics such as F1 and recall, suggesting that different modeling strategies favor different operating regimes. Third, performance drops markedly in the open-world setting, highlighting the challenge of recognizing previously unseen abnormalities.

\subsection{Robustness to test-time perturbations}

\begin{figure}[t]
    \centering
    \includegraphics[width=\linewidth]{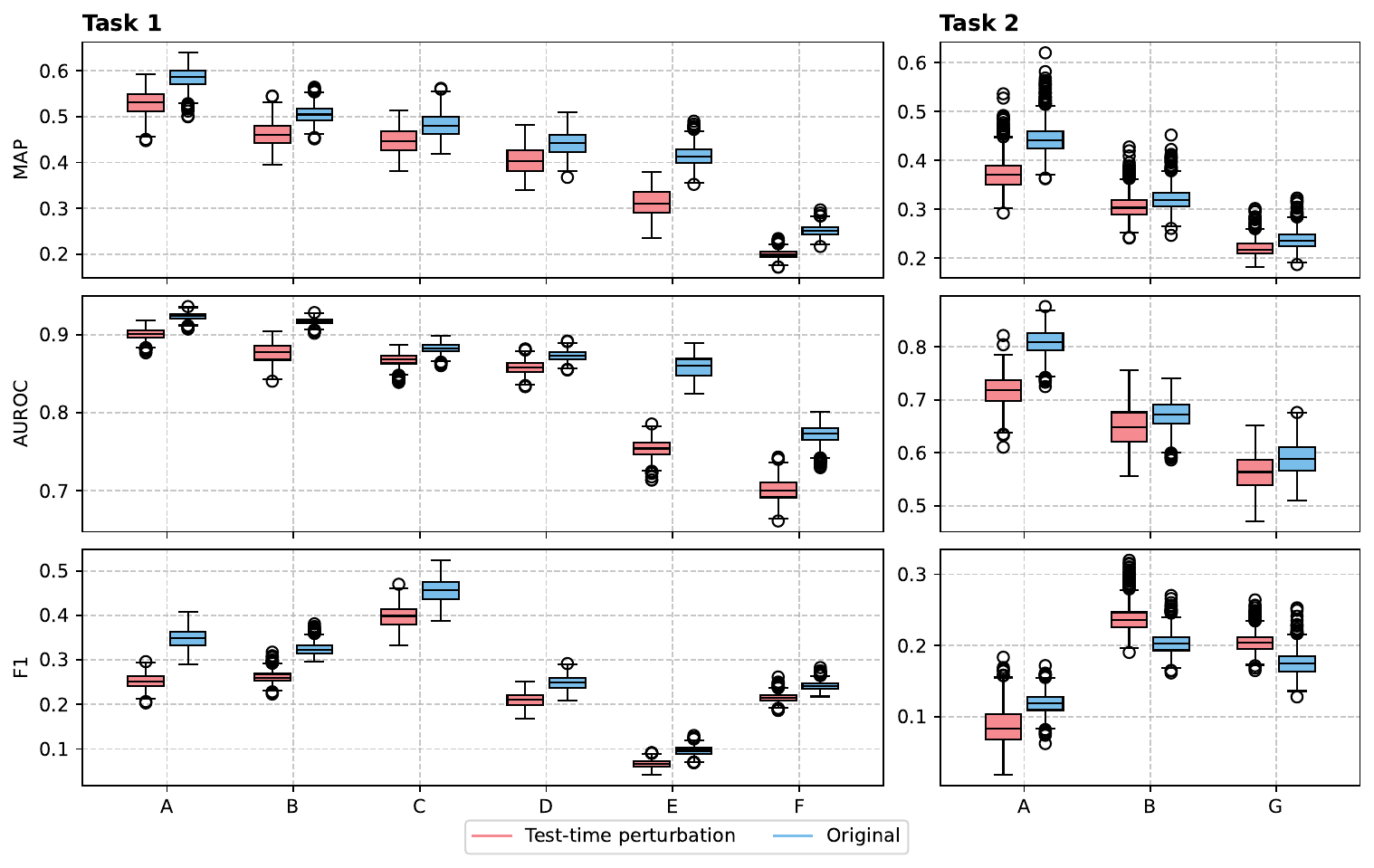}
    \caption{
    Robustness analysis under test-time perturbations. 
    Each panel compares model performance under test-time perturbations and original test data. 
    }
    \label{fig:robustness_fig}
\end{figure}

To evaluate the robustness of the models, we conducted an additional analysis under test-time perturbations. Specifically, we applied augmentation-based transformations to the test images and compared model performance against the original evaluation setting. This analysis allows us to examine how sensitive model predictions are to small variations in the input data. \Cref{fig:robustness_fig} illustrates the performance differences across teams and evaluation metrics, and \Cref{tab:task1_aug_ci} summarizes the quantitative results for Task 1 and Task 2.
Overall, most methods exhibit performance decreases under test-time perturbations. In Task 1, the best-performing method achieves an mAP of 0.5854 under the original setting but drops to 0.5292 with perturbations. Similar declines are observed across other teams, suggesting that even high-performing models remain sensitive to input variations.

The effect is more pronounced for ranking-based metrics (e.g., mAP and AUROC), whereas threshold-dependent metrics (e.g., F1) show greater variability across teams. This observation indicates that perturbations mainly affect the relative ranking rather than fixed decision thresholds. A comparable trend is observed in Task 2. Although baseline performance is lower due to the open-world setting, test-time perturbations still lead to consistent drops across most teams. 
\begin{figure}
    \centering
\includegraphics[width=\linewidth]{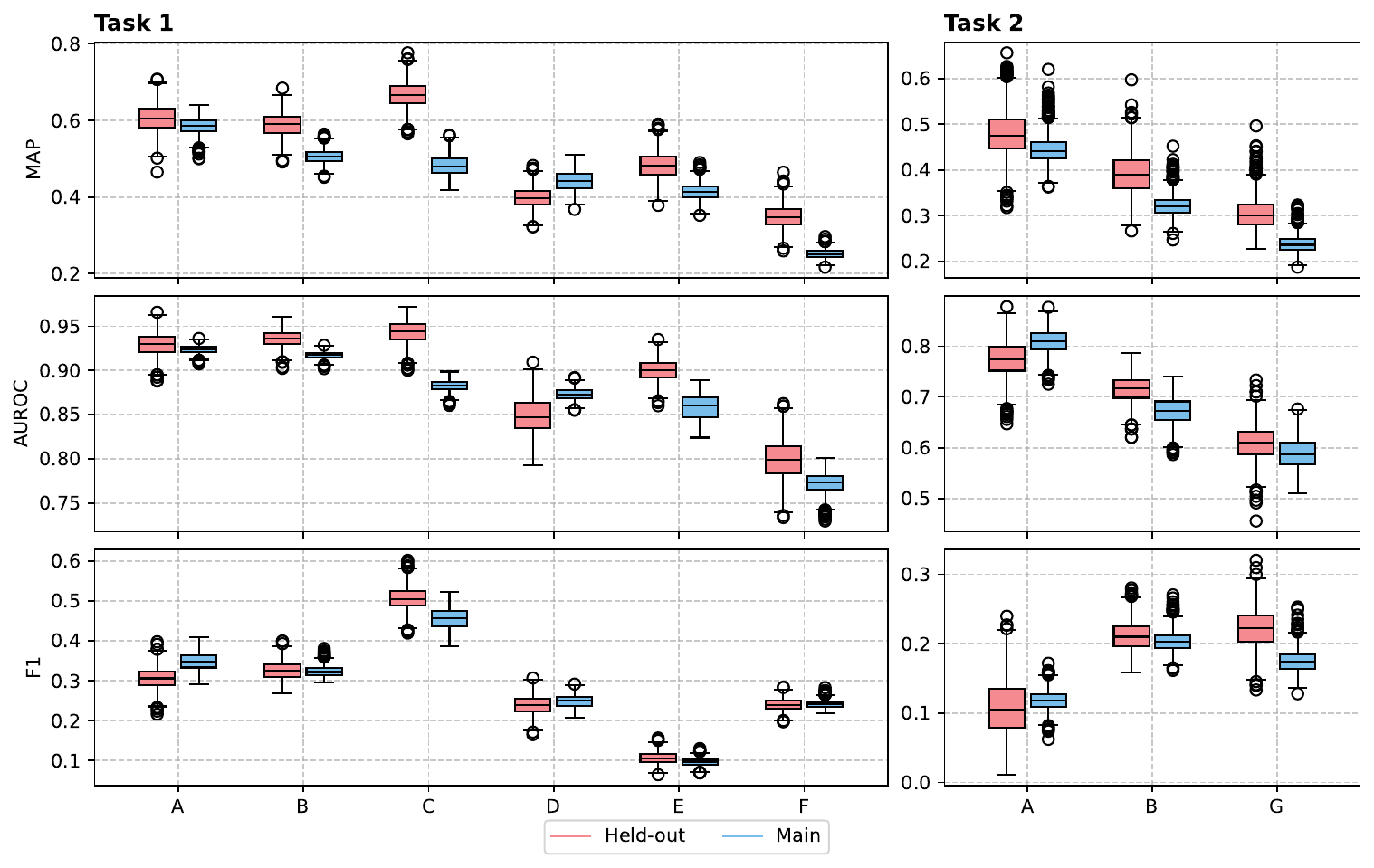}
\caption{Evaluation on held-out subsets for Task 1 and Task 2. 
Each panel compares model performance across teams under held-out subsets and the main evaluation. 
The left column shows results for Task 1, and the right column shows results for Task 2, with mAP, AUROC, and F1 reported as representative metrics. 
Across most teams, the relative ranking of methods remains consistent between evaluation settings, while overall performance tends to decrease on the held-out subsets.
}
\label{fig:cheat_fig}
\end{figure}

\subsection{Evaluation on held-out subsets}

To further examine the stability of model performance, we conducted additional evaluations on held-out subsets constructed from the available data. For Task 1, a subset of 300 cases was sampled from the training data, while for Task 2, a subset of 200 cases was selected from the development set. All models were evaluated on these subsets using the same protocol as in the main evaluation.

The results are summarized in \Cref{fig:cheat_fig} and detailed in \Cref{tab:task1_cheat_ci_2rows}. Overall, the relative performance of different methods remains largely consistent with the main evaluation. Methods that perform well on the full evaluation set tend to achieve higher scores on the held-out subsets as well, indicating that the ranking of methods is generally stable across different data samples. At the same time, performance on the held-out subsets is typically lower than on the main evaluation set. This trend is consistent across most teams and metrics and may be attributed to differences in data composition and the smaller sample size of the subsets.

We also observe that the impact varies across metrics. Ranking-based metrics such as mAP and AUROC show more consistent trends across evaluation settings, while threshold-dependent metrics such as F1 exhibit greater variability. This suggests that model ranking is relatively robust, whereas threshold-based performance is more sensitive to changes in evaluation data. 

\subsection{Head-to-tail performance}

\begin{figure}
\centering
    \begin{subfigure}[t]{0.63\textwidth}
        \centering
        \includegraphics[width=\textwidth]{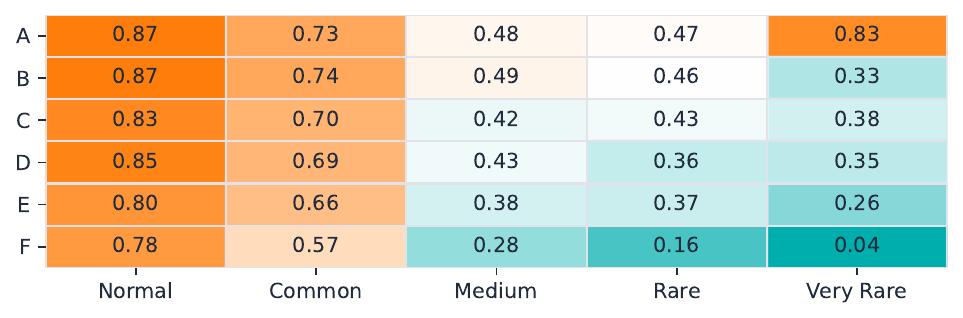}
       
        \caption{Head-to-tail performance on Task 1}
        \label{fig:generalization_analysis_a}
    \end{subfigure}
    \hfill
    \begin{subfigure}[t]{0.33\textwidth}
        \centering
         \includegraphics[width=\textwidth]{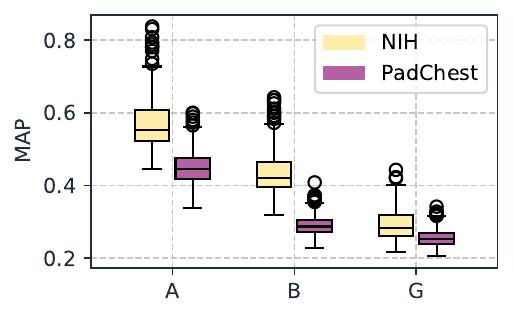}
       
        \caption{Cross-center evaluation on Task 2}
        \label{fig:generalization_analysis_b}
    \end{subfigure}

\caption{
Generalization analysis across disease frequency and clinical centers.
a) Head-to-tail performance on Task 1 across disease groups with different occurrence frequencies.
b) Cross-center evaluation on Task 2, comparing performance between the NIH ChestX-ray and PadChest subsets.
}
\label{fig:generalization_analysis}
\end{figure}

To better understand model behavior under long-tailed disease distributions, we analyzed performance across disease frequency. Diseases were grouped into five categories: normal cases, common diseases ($>10\%$), medium-frequency diseases ($1\%-10\%$), rare diseases ($0.1\% - 1\%$), and very rare diseases ($<0.1\%$). As shown in \Cref{fig:generalization_analysis_a} and detailed in \Cref{tab:task1_rare_results}, model performance consistently decreases as disease frequency becomes lower. While most methods achieve strong performance on normal and common disease categories, their performance drops noticeably for rare and very rare diseases. This trend highlights the persistent challenge of long-tailed recognition in CXR analysis. In particular, the performance gap between common and rare disease groups remains substantial across all teams, suggesting that current methods still struggle to learn and generalize visual patterns for low-frequency abnormalities.

\subsection{Multi-center generalization}

To further examine cross-center generalization, we analyzed model
performance separately on the PadChest-GR internal test set and the
NIH ChestX-ray external test set for Task~2. As shown in \Cref{fig:generalization_analysis_b} and detailed in \Cref{tab:task2_multicenter_ci}, all participating methods achieved
higher mAP on the NIH subset than on the PadChest-GR subset.

To characterize the heterogeneity between the two evaluation sources, we further compared their label composition, acquisition metadata, and
image characteristics, as summarized in \Cref{tab:intercenter_heterogeneity}. The PadChest-GR internal test
set contained 305 images and was dominated by scoliosis
(245/305, 80.3\%). In contrast, the NIH external test set contained 79
images with a more distributed label composition, including scoliosis
(28/79, 35.4\%), bulla (21/79, 26.6\%), adenopathy
(15/79, 19.0\%), goiter (11/79, 13.9\%), infarction
(3/79, 3.8\%), and osteopenia (1/79, 1.3\%).

The two subsets also differed substantially in image acquisition and
representation. PadChest-GR images had a higher and more variable
native resolution, with an average size of
$2690.9 \times 2534.8$ pixels and an average resolution of
$7.09$ megapixels. By comparison, the NIH subset consisted of
$1024 \times 1024$ images with an average resolution of
$1.05$ megapixels. After normalization to $[0,1]$, the NIH images had
a higher mean grayscale intensity than the PadChest-GR images
($0.450$ vs.\ $0.397$) and a lower grayscale intensity standard
deviation ($0.246$ vs.\ $0.282$). The subsets also differed in
view-position composition: AP images accounted for 17.7\% of the NIH
subset but only 1.6\% of the PadChest-GR subset. Formal image-quality
ratings were not available for either source; therefore, we report
resolution and intensity statistics as directly measurable image
characteristics rather than formal measures of diagnostic image
quality.

These findings show that the two subsets differ not only by institution,
but also by class prevalence, case composition, view position,
resolution, and intensity distribution. The observed performance gap
should therefore not be attributed solely to center identity. Although
NIH ChestX-ray represents an external source, the NIH subset used in
CXR-LT 2026 was neither a random nor an unfiltered external cohort. The original
NIH ChestX-ray dataset was collected at the NIH Clinical Center, a
clinical research setting that may differ from routine hospital cohorts
in case spectrum~\citep{wang2017chestxray8,gallin2011nih}. In addition,
the CXR-LT 2026 NIH subset was enriched through report-based candidate
screening followed by radiologist confirmation and adjudication. This
process may have retained more clinically evident or prototypical
examples of the six target findings while excluding ambiguous
candidates.

Several other factors may also contribute to the higher NIH mAP. Average precision is sensitive to class prevalence and the positive and negative composition of the evaluation set, particularly in small, long-tailed multi-label datasets. The annotation pipelines also differed:
PadChest-GR labels were mapped from the original PadChest-GR taxonomy,
whereas the NIH cases were reviewed directly using the six Task~2 label
definitions. Given these differences and the small size of the NIH
external test set, the comparison should be interpreted
as an exploratory cross-center analysis conditioned on both label and
image heterogeneity, rather than as evidence that external-source
evaluation is intrinsically easier or that cross-center generalization
has been fully resolved.

\section{Discussion}

In this study, we present the CXR-LT 2026 challenge, a multi-center benchmark designed to evaluate long-tailed learning and open-world generalization in chest X-ray analysis. Compared with previous editions, this benchmark introduces radiologist-annotated evaluation sets and cross-center testing, providing a more realistic setting for assessing model performance.

Several key observations emerge from the results. First, although top-performing methods achieve strong ranking performance on common findings, substantial performance gaps remain across teams, particularly under long-tailed distributions. Performance consistently decreases for low-frequency disease categories, indicating that current methods still struggle to capture visual patterns for rare abnormalities. This gap suggests that existing strategies for handling class imbalance, such as reweighting or resampling, remain insufficient.

Second, we observe a clear mismatch between different evaluation metrics. As shown in \Cref{fig:radar_task1}, ranking-based metrics do not always align with threshold-dependent metrics. Methods that perform well in mAP or AUROC do not necessarily achieve strong classification performance in F1 or recall, suggesting differences in prediction calibration and optimal thresholds. Consistently, many methods also exhibit suboptimal calibration, indicating that predicted probabilities are not always reliable for clinical decision support.

Third, performance drops significantly in the open-world setting (\Cref{fig:radar_task2}), where models are required to recognize unseen disease categories. While vision-language models improve zero-shot performance, the overall results suggest that generalization to new disease concepts remains limited. This finding highlights the gap between current supervised learning approaches and real-world clinical requirements.

Beyond these main findings, additional analyses provide further insights into model behavior. The robustness experiments show that model predictions remain sensitive to input perturbations (\Cref{fig:robustness_fig}), even for top-performing methods. Evaluation on held-out subsets demonstrates that while relative rankings are generally stable, absolute performance can vary across data splits. Multi-center analysis further reveals that external-source evaluation does not necessarily imply lower measured performance. In Task~2, mAP was higher on the NIH ChestX-ray subset than on the PadChest-GR subset for all evaluated methods. As shown by the descriptive comparison in \Cref{tab:intercenter_heterogeneity}, the two evaluation subsets
also differed substantially in label composition, view-position
distribution, image resolution, and intensity characteristics. This reflects not only institutional domain shift but also differences in the case spectrum, disease conspicuity, class prevalence, annotation pipeline, and sample size. In particular, the NIH subset may contain more clinically evident or prototypical rare findings because it was derived from a clinical research hospital dataset and further enriched through report screening and radiologist confirmation. These results therefore motivate larger external evaluations with broader case spectra and better-matched class distributions.

The use of external pretrained models should also be considered when interpreting the leaderboard. CXR-LT 2026 allowed external data and publicly available pretrained models under the non-overlap rules described in \Cref{sec:Evaluation}, because such resources are now common in state-of-the-art radiology AI systems. This policy makes the challenge more representative of contemporary model development, but it also means that the leaderboard compares complete submitted systems rather than isolated downstream learning algorithms. Performance may therefore reflect not only model architecture and loss design, but also the scale and domain relevance of external pretraining, prompt engineering, fine-tuning choices, and ensembling. For Task 2, we mitigated this concern by requiring the six OOD labels to remain unseen during supervised training and by prohibiting models pretrained using images or similar label supervision for these OOD categories.

This benchmark has several limitations. First, although the NIH ChestX-ray subset provides an external-source evaluation for Task 2, it is relatively small, containing 79 radiologist-confirmed cases, and was intentionally enriched for rare Task 2 findings. Therefore, it should be interpreted as an initial cross-center stress test rather than a definitive estimate of real-world external performance. Larger external cohorts with broader case spectra, matched class distributions, and standardized annotation protocols will be needed to more fully characterize cross-center generalization. Second, the large-scale PadChest training set relies on report-derived weak labels. This design enables scalable model development under realistic weak-supervision conditions, but the labels may contain noise due to report parsing, uncertain or negated findings, synonym mapping, and imperfect alignment between report-level descriptions and image-level abnormalities. Third, CXR-LT 2026 allowed external data and pretrained models under the non-overlap and Task 2 OOD-label restrictions described in \Cref{sec:Evaluation}. This policy reflects contemporary radiology AI practice, but it also means that leaderboard performance reflects complete submitted systems, including external pretraining, prompt design, fine-tuning strategy, and ensembling, rather than isolated downstream architectures alone. These limitations should be considered when interpreting the benchmark results and motivate future challenge editions with larger external test sets, more detailed data heterogeneity analyses, and continued efforts to reduce label noise.

\section{Conclusion}

The CXR-LT 2026 benchmark highlights several open challenges for radiology AI, including long-tailed recognition, probability calibration, robustness to data variation, and cross-center generalization. Addressing these issues will likely require improved representation learning, better calibration methods, and approaches that explicitly account for distribution shifts across clinical settings. The benchmark provides a unified testbed for studying these problems under realistic conditions and for developing more robust and clinically applicable models.

.

\section*{CRediT authorship contribution statement}

Visualization: Y.L., H.D., Y.P.
Formal analysis: H.D., Y.L.;
Writing – review \& editing: H.D., Y.L., P.Z., F.Z., A.C.L., J.C., D.K., J.N.K., M.K., S.K., G.M., K.T.N., T.H,N., Ha.H.P., Huy.H.P., Huy.L.P., N.R.S., A.T., R.Z., A.Z., A.E.F., Z.L., R.M.S., M.L., H.C., Y.Y., G.S., Y.P.;
Validation: H.D., Y.L.;
Data curation: H.D., Y.L., P.Z., F.Z., A.C.L.;
Writing – original draft: H.D., Y.L.;
Methodology: H.D., Y.L., Y.P.;
Resources: Y.P.;
Software: G.S.;
Supervision: Y.P.;
Conceptualization: H.D., Y.L., Y.P.;
Project administration: Y.P.;
Funding acquisition: Y.P.

\section*{Acknowledgments}
\label{sec:acknowledgments}

This work was supported by the National Library of Medicine [grant number R01LM014306], the National Science Foundation (NSF) [grant numbers 2145640], and the Cornell–HKUST Global Strategic Collaboration Award. It was also supported by the NIH Intramural Research Program, the National Library of Medicine and Clinical Center.

\section*{Disclaimer}
The contributions of the NIH author(s) are considered Works of the United States Government. The findings and conclusions presented in this paper are those of the author(s) and do not necessarily reflect the views of the NIH or the U.S. Department of Health and Human Services.

\section*{Conflict of interest statement}
Declaration of competing interest The authors declare the following financial interests/personal relationships which may be considered potential competing interests: R.M.S. has received royalties for patent or software licenses from iCAD, Philips, PingAn, ScanMed, Translation Holdings, and MGB, as well as research support from a CRADA with PingAn. The remaining authors declare that they have no known competing financial interests or personal relationships that could have appeared to influence the work reported in this paper.

\bibliographystyle{unsrtnat}

\bibliography{cas-refs}


\appendix

\renewcommand{\tablename}{Extended Data Table}
\clearpage

\renewcommand{\tablename}{Extended Data Table}
\crefalias{table}{extendedtable}
\setcounter{table}{0}

\newcommand{\twolineci}[3]{#1\,(#2-#3)}
\newcommand{\cmark}{\ding{51}}%
\newcommand{\xmark}{\ding{55}}%

\newcolumntype{Y}{>{\raggedright\arraybackslash}X}


\section{Supplementary materials}

\begin{table}[H]
\centering
\tiny
\setlength{\tabcolsep}{2pt}
\caption{Summary of annotation protocols and agreement/adjudication flow for the CXR-LT 2026 evaluation sets.}
\label{tab:annotation_protocol}
\begin{tabularx}{\linewidth}{
    >{\hsize=.7\hsize}Y
    l
    >{\hsize=1\hsize}Y
    >{\hsize=1\hsize}Y
    >{\hsize=1\hsize}Y
    >{\hsize=1\hsize}Y
    >{\hsize=1.6\hsize}Y
    >{\hsize=.7\hsize}Y
}
\toprule
\textbf{Evaluation set} &
\textbf{Source} &
\textbf{Candidate selection / label-space construction} &
\textbf{Annotation source or platform} &
\textbf{Readers} &
\textbf{Annotation instructions} &
\textbf{Disagreement handling and agreement summary} &
\textbf{Final evaluation set} \\
\midrule

Task~1 development and test sets &
PadChest-GR &
One radiologist (P.Z.) reviewed the PadChest-GR finding categories and curated the mapping from the original PadChest-GR labels to the CXR-LT 2026 challenge taxonomy.  &
Radiologist-reviewed PadChest-GR annotations; CXR-LT label mapping curated from the released PadChest-GR label files. &
Original PadChest-GR radiologist annotators; CXR-LT mapping curated by P.Z. &
Original PadChest-GR findings were mapped to CXR-LT 2026 image-level categories. &
CXR-LT 2026 did not generate new image-level annotations for these cases; instead, it relied on the radiologist-reviewed PadChest-GR labels and the original PadChest-GR quality-control/adjudication protocol. &
1,620 development cases and 1,400 test cases. \\

\midrule

Task~2 development and internal test sets &
PadChest-GR &
The Task~2 label set was defined from the same radiologist-curated mapping process. &
Radiologist-reviewed PadChest-GR annotations; CXR-LT label mapping curated from the released PadChest-GR label files. &
Original PadChest-GR radiologist annotators; CXR-LT mapping curated by P.Z. &
Original PadChest-GR findings were mapped to the six Task~2 image-level out-of-distribution categories. &
CXR-LT 2026 relied on the radiologist-reviewed PadChest-GR labels and the original PadChest-GR quality-control/adjudication protocol. &
200 development cases and 305 internal test cases. \\

\midrule

Task~2 external test set &
NIH ChestX-ray &
159 CXR cases were selected as candiate images and final labels were determined by radiologist image review. &
MD.ai annotation platform. &
First-stage independent review by P.Z.[5 years of experience] and F.Z.[5 years of experience]; discordant cases adjudicated by A.C.L [21 years of experience]. &
Readers were provided with the six Task~2 label definitions and assigned image-level binary labels indicating the presence or absence of the target finding(s). &
P.Z. and F.Z. independently reviewed all 159 candidates. 30 cases were confirmed by both readers and retained directly. 53 cases had discordant initial interpretations, defined as one reader assigning positive and the other assigning negative, and were reviewed by A.C.L.; 49 of these discordant cases were confirmed and retained. The remaining candidates were excluded from the final external test set. &
79 external test cases: 30 two-reader concordant positives plus 49 adjudicated positives. \\

\bottomrule
\end{tabularx}
\end{table}


\clearpage

\begin{table}[htbp]
\centering
\caption{Main results with 95\% bootstrap confidence intervals based on 1,000 resampling runs.}
\label{tab:task1_main_ci}
\footnotesize
\setlength{\tabcolsep}{4pt}
\begin{tabular}{l *{6}{c}}
\toprule
Team & mAP & AUROC & F1 & ECE & Precision & Recall \\
\midrule
\rowcolor{GreenYellow}\multicolumn{7}{l}{Task 1}\\
A &
\twolineci{0.585}{0.540}{0.621} &
\twolineci{0.926}{0.915}{0.932} &
\twolineci{0.352}{0.305}{0.390} &
\twolineci{0.920}{0.912}{0.923} &
\twolineci{0.590}{0.498}{0.618} &
\twolineci{0.296}{0.255}{0.337} \\
B &
\twolineci{0.483}{0.473}{0.543} &
\twolineci{0.919}{0.909}{0.925} &
\twolineci{0.316}{0.303}{0.354} &
\twolineci{0.918}{0.910}{0.921} &
\twolineci{0.531}{0.472}{0.596} &
\twolineci{0.267}{0.255}{0.300} \\
C &
\twolineci{0.460}{0.437}{0.532} &
\twolineci{0.883}{0.870}{0.894} &
\twolineci{0.450}{0.405}{0.499} &
\twolineci{0.801}{0.784}{0.804} &
\twolineci{0.439}{0.383}{0.488} &
\twolineci{0.505}{0.456}{0.574} \\
D &
\twolineci{0.430}{0.396}{0.487} &
\twolineci{0.875}{0.861}{0.886} &
\twolineci{0.248}{0.218}{0.275} &
\twolineci{0.916}{0.908}{0.919} &
\twolineci{0.450}{0.372}{0.481} &
\twolineci{0.207}{0.187}{0.237} \\
E &
\twolineci{0.395}{0.373}{0.458} &
\twolineci{0.859}{0.834}{0.881} &
\twolineci{0.095}{0.077}{0.115} &
\twolineci{0.612}{0.603}{0.614} &
\twolineci{0.368}{0.311}{0.409} &
\twolineci{0.061}{0.047}{0.077} \\
F &
\twolineci{0.236}{0.231}{0.276} &
\twolineci{0.776}{0.746}{0.792} &
\twolineci{0.235}{0.225}{0.262} &
\twolineci{0.726}{0.702}{0.738} &
\twolineci{0.168}{0.161}{0.189} &
\twolineci{0.520}{0.496}{0.569} \\

\multicolumn{7}{l}{\cellcolor{GreenYellow}{Task 2}}\\
A &
\twolineci{0.432}{0.390}{0.523} &
\twolineci{0.807}{0.755}{0.850} &
\twolineci{0.118}{0.091}{0.147} &
\twolineci{0.452}{0.433}{0.464} &
\twolineci{0.267}{0.233}{0.322} &
\twolineci{0.102}{0.072}{0.133} \\
B &
\twolineci{0.311}{0.278}{0.380} &
\twolineci{0.671}{0.614}{0.723} &
\twolineci{0.203}{0.178}{0.237} &
\twolineci{0.175}{0.153}{0.199} &
\twolineci{0.196}{0.174}{0.233} &
\twolineci{0.486}{0.394}{0.608} \\
G &
\twolineci{0.224}{0.204}{0.278} &
\twolineci{0.590}{0.538}{0.657} &
\twolineci{0.174}{0.147}{0.206} &
\twolineci{0.573}{0.525}{0.590} &
\twolineci{0.182}{0.157}{0.221} &
\twolineci{0.385}{0.325}{0.464} \\
\bottomrule
\end{tabular}
\end{table}

\clearpage

\begin{table}[!hptb]
\centering
\caption{Augmentation results with 95\% bootstrap confidence intervals based on 1,000 resampling runs.}
\label{tab:task1_aug_ci}
\footnotesize
\setlength{\tabcolsep}{4pt}

\begin{tabular}{l *{6}{c}}
\toprule
Team & mAP & AUROC & F1 & ECE & Precision & Recall \\
\midrule
\multicolumn{7}{l}{\cellcolor{GreenYellow}{Task 1}}\\
A &
\twolineci{0.529}{0.469}{0.579} &
\twolineci{0.903}{0.886}{0.913} &
\twolineci{0.254}{0.220}{0.280} &
\twolineci{0.920}{0.912}{0.923} &
\twolineci{0.522}{0.441}{0.545} &
\twolineci{0.194}{0.170}{0.227} \\
B &
\twolineci{0.443}{0.414}{0.513} &
\twolineci{0.878}{0.854}{0.897} &
\twolineci{0.256}{0.239}{0.289} &
\twolineci{0.919}{0.912}{0.922} &
\twolineci{0.564}{0.477}{0.619} &
\twolineci{0.202}{0.188}{0.232} \\
C &
\twolineci{0.433}{0.399}{0.499} &
\twolineci{0.870}{0.849}{0.883} &
\twolineci{0.394}{0.354}{0.443} &
\twolineci{0.772}{0.754}{0.777} &
\twolineci{0.367}{0.328}{0.420} &
\twolineci{0.458}{0.408}{0.524} \\
D &
\twolineci{0.393}{0.354}{0.454} &
\twolineci{0.860}{0.841}{0.876} &
\twolineci{0.212}{0.181}{0.238} &
\twolineci{0.916}{0.908}{0.918} &
\twolineci{0.459}{0.389}{0.489} &
\twolineci{0.159}{0.136}{0.189} \\
E &
\twolineci{0.302}{0.263}{0.365} &
\twolineci{0.756}{0.731}{0.774} &
\twolineci{0.065}{0.050}{0.083} &
\twolineci{0.608}{0.598}{0.609} &
\twolineci{0.349}{0.274}{0.381} &
\twolineci{0.040}{0.030}{0.053} \\
F &
\twolineci{0.188}{0.183}{0.219} &
\twolineci{0.698}{0.674}{0.728} &
\twolineci{0.210}{0.199}{0.235} &
\twolineci{0.753}{0.730}{0.764} &
\twolineci{0.154}{0.146}{0.173} &
\twolineci{0.387}{0.356}{0.433} \\
\multicolumn{7}{l}{\cellcolor{GreenYellow}{Task 2}}\\
A &
\twolineci{0.360}{0.319}{0.447} &
\twolineci{0.716}{0.656}{0.768} &
\twolineci{0.087}{0.042}{0.139} &
\twolineci{0.457}{0.440}{0.470} &
\twolineci{0.359}{0.111}{0.429} &
\twolineci{0.054}{0.024}{0.091} \\
B &
\twolineci{0.295}{0.266}{0.363} &
\twolineci{0.649}{0.582}{0.732} &
\twolineci{0.236}{0.206}{0.284} &
\twolineci{0.167}{0.145}{0.187} &
\twolineci{0.219}{0.193}{0.271} &
\twolineci{0.536}{0.436}{0.664} \\
G &
\twolineci{0.206}{0.191}{0.262} &
\twolineci{0.563}{0.495}{0.624} &
\twolineci{0.203}{0.181}{0.234} &
\twolineci{0.566}{0.527}{0.582} &
\twolineci{0.174}{0.161}{0.205} &
\twolineci{0.556}{0.453}{0.688} \\
\bottomrule
\end{tabular}
\end{table}

\clearpage

\begin{table}[H]
\centering
\caption{Evaluation results on held-out subsets with 95\% bootstrap confidence intervals based on 1,000 resampling runs. For each team, we report results trained with and without augmentation.}
\label{tab:task1_cheat_ci_2rows}
\scriptsize
\setlength{\tabcolsep}{4pt}
\renewcommand{\arraystretch}{1.15}

\begin{tabular}{lcc c c c c c}
\toprule
Team & Aug & mAP & mAUC & mF1 & mECE & Precision & Recall \\
\midrule
\multicolumn{8}{l}{\cellcolor{GreenYellow}{Task 1}}\\
{A} & \xmark &
\twolineci{0.591}{0.539}{0.674} &
\twolineci{0.928}{0.903}{0.953} &
\twolineci{0.313}{0.252}{0.361} &
\twolineci{0.925}{0.916}{0.930} &
\twolineci{0.475}{0.367}{0.545} &
\twolineci{0.278}{0.228}{0.326} \\
& \cmark &
\twolineci{0.458}{0.430}{0.548} &
\twolineci{0.894}{0.872}{0.919} &
\twolineci{0.260}{0.196}{0.300} &
\twolineci{0.924}{0.914}{0.929} &
\twolineci{0.417}{0.317}{0.467} &
\twolineci{0.220}{0.164}{0.262} \\
\rowcolor{gray!20}
{B} & \xmark &
\twolineci{0.551}{0.532}{0.650} &
\twolineci{0.935}{0.917}{0.953} &
\twolineci{0.316}{0.282}{0.375} &
\twolineci{0.925}{0.916}{0.929} &
\twolineci{0.487}{0.394}{0.559} &
\twolineci{0.272}{0.252}{0.327} \\
\rowcolor{gray!20}
& \cmark &
\twolineci{0.420}{0.403}{0.523} &
\twolineci{0.900}{0.872}{0.921} &
\twolineci{0.233}{0.192}{0.287} &
\twolineci{0.924}{0.914}{0.929} &
\twolineci{0.453}{0.305}{0.511} &
\twolineci{0.189}{0.157}{0.243} \\
{C} & \xmark &
\twolineci{0.652}{0.602}{0.732} &
\twolineci{0.945}{0.920}{0.965} &
\twolineci{0.498}{0.449}{0.563} &
\twolineci{0.806}{0.790}{0.812} &
\twolineci{0.419}{0.374}{0.494} &
\twolineci{0.698}{0.637}{0.790} \\
& \cmark &
\twolineci{0.506}{0.472}{0.598} &
\twolineci{0.912}{0.881}{0.934} &
\twolineci{0.402}{0.355}{0.462} &
\twolineci{0.771}{0.753}{0.780} &
\twolineci{0.326}{0.286}{0.392} &
\twolineci{0.585}{0.524}{0.674} \\
\rowcolor{gray!20}
{D} & \xmark &
\twolineci{0.375}{0.348}{0.451} &
\twolineci{0.838}{0.811}{0.889} &
\twolineci{0.245}{0.191}{0.281} &
\twolineci{0.917}{0.908}{0.922} &
\twolineci{0.345}{0.271}{0.403} &
\twolineci{0.218}{0.170}{0.254} \\
\rowcolor{gray!20}
& \cmark &
\twolineci{0.340}{0.311}{0.410} &
\twolineci{0.832}{0.805}{0.873} &
\twolineci{0.219}{0.166}{0.252} &
\twolineci{0.916}{0.906}{0.921} &
\twolineci{0.373}{0.268}{0.417} &
\twolineci{0.185}{0.136}{0.219} \\
{E} & \cmark &
\twolineci{0.455}{0.415}{0.548} &
\twolineci{0.900}{0.876}{0.923} &
\twolineci{0.105}{0.079}{0.136} &
\twolineci{0.615}{0.606}{0.618} &
\twolineci{0.283}{0.181}{0.323} &
\twolineci{0.081}{0.066}{0.105} \\
& \cmark &
\twolineci{0.303}{0.268}{0.387} &
\twolineci{0.800}{0.767}{0.836} &
\twolineci{0.064}{0.034}{0.093} &
\twolineci{0.611}{0.601}{0.614} &
\twolineci{0.240}{0.115}{0.278} &
\twolineci{0.043}{0.020}{0.068} \\
\rowcolor{gray!20}
{F} & \xmark &
\twolineci{0.322}{0.293}{0.406} &
\twolineci{0.798}{0.756}{0.842} &
\twolineci{0.231}{0.212}{0.271} &
\twolineci{0.698}{0.667}{0.726} &
\twolineci{0.155}{0.142}{0.185} &
\twolineci{0.688}{0.620}{0.757} \\
\rowcolor{gray!20}
& \cmark &
\twolineci{0.223}{0.207}{0.303} &
\twolineci{0.737}{0.688}{0.793} &
\twolineci{0.207}{0.178}{0.246} &
\twolineci{0.705}{0.670}{0.732} &
\twolineci{0.136}{0.118}{0.169} &
\twolineci{0.542}{0.462}{0.630} \\
\midrule
\multicolumn{8}{l}{\cellcolor{GreenYellow}{Task 2}}\\
{A} & \xmark &
\twolineci{0.460}{0.376}{0.593} &
\twolineci{0.776}{0.702}{0.838} &
\twolineci{0.114}{0.037}{0.192} &
\twolineci{0.448}{0.428}{0.468} &
\twolineci{0.442}{0.189}{0.500} &
\twolineci{0.081}{0.026}{0.149} \\
& \cmark &
\twolineci{0.333}{0.283}{0.451} &
\twolineci{0.716}{0.638}{0.781} &
\twolineci{0.000}{0.000}{0.000} &
\twolineci{0.453}{0.432}{0.475} &
\twolineci{0.000}{0.000}{0.000} &
\twolineci{0.000}{0.000}{0.000} \\
\rowcolor{gray!20}
{B} & \xmark &
\twolineci{0.375}{0.305}{0.487} &
\twolineci{0.716}{0.665}{0.768} &
\twolineci{0.213}{0.171}{0.257} &
\twolineci{0.178}{0.144}{0.216} &
\twolineci{0.224}{0.193}{0.259} &
\twolineci{0.562}{0.442}{0.663} \\
\rowcolor{gray!20}
& \cmark &
\twolineci{0.361}{0.294}{0.468} &
\twolineci{0.684}{0.617}{0.742} &
\twolineci{0.233}{0.186}{0.280} &
\twolineci{0.148}{0.123}{0.184} &
\twolineci{0.233}{0.197}{0.277} &
\twolineci{0.549}{0.424}{0.661} \\
{G} & \xmark &
\twolineci{0.280}{0.248}{0.385} &
\twolineci{0.608}{0.542}{0.674} &
\twolineci{0.226}{0.165}{0.276} &
\twolineci{0.550}{0.535}{0.578} &
\twolineci{0.259}{0.214}{0.307} &
\twolineci{0.455}{0.321}{0.585} \\
& \cmark &
\twolineci{0.248}{0.226}{0.329} &
\twolineci{0.574}{0.510}{0.641} &
\twolineci{0.253}{0.214}{0.287} &
\twolineci{0.527}{0.509}{0.554} &
\twolineci{0.229}{0.207}{0.253} &
\twolineci{0.660}{0.525}{0.784} \\
\bottomrule
\end{tabular}
\end{table}

\clearpage

\begin{table}[H]
\caption{Head to Tail Results in Task 1.}
\label{tab:task1_rare_results}
\centering
\begin{tabular}{lccccc}
\toprule
Team & Normal & Common & Medium & Rare & Very Rare \\
\midrule
A & 0.873 & 0.734 & 0.484 & 0.468 & 0.826 \\
B & 0.870 & 0.736 & 0.489 & 0.455 & 0.328 \\
C & 0.835 & 0.695 & 0.424 & 0.435 & 0.381 \\
D & 0.852 & 0.690 & 0.429 & 0.359 & 0.347 \\
E & 0.796 & 0.661 & 0.384 & 0.368 & 0.258 \\
F & 0.780 & 0.568 & 0.281 & 0.157 & 0.038 \\
\bottomrule
\end{tabular}
\end{table}

\clearpage

\begin{table}[H]
\centering
\caption{Multi-center evaluation with 95\% bootstrap confidence intervals (1,000 runs) in Task~2. Results are reported separately on PadChest and NIH subsets.}
\label{tab:task2_multicenter_ci}
\footnotesize
\setlength{\tabcolsep}{4pt}

\begin{tabular}{l c c c c c c}
\toprule
Team & mAP & mAUC & mF1 & mECE & Precision & Recall \\
\midrule
\multicolumn{7}{l}{\cellcolor{GreenYellow}{PadChest}}\\
{A} & 
\twolineci{0.436}{0.371}{0.538} &
\twolineci{0.767}{0.718}{0.815} &
\twolineci{0.130}{0.093}{0.164} &
\twolineci{0.450}{0.434}{0.467} &
\twolineci{0.313}{0.274}{0.354} &
\twolineci{0.115}{0.075}{0.155} \\
{B} & 
\twolineci{0.272}{0.246}{0.340} &
\twolineci{0.647}{0.591}{0.704} &
\twolineci{0.193}{0.162}{0.225} &
\twolineci{0.146}{0.126}{0.173} &
\twolineci{0.217}{0.194}{0.253} &
\twolineci{0.458}{0.368}{0.550} \\
{G} & 
\twolineci{0.241}{0.219}{0.308} &
\twolineci{0.560}{0.512}{0.605} &
\twolineci{0.189}{0.146}{0.227} &
\twolineci{0.524}{0.511}{0.547} &
\twolineci{0.217}{0.188}{0.250} &
\twolineci{0.410}{0.304}{0.510} \\
\midrule
\multicolumn{7}{l}{\cellcolor{GreenYellow}{NIH ChestX-ray}}\\
{A} & 
\twolineci{0.514}{0.486}{0.673} &
\twolineci{0.840}{0.786}{0.892} &
\twolineci{0.141}{0.095}{0.205} &
\twolineci{0.408}{0.365}{0.430} &
\twolineci{0.250}{0.167}{0.400} &
\twolineci{0.112}{0.061}{0.180} \\
{B} & 
\twolineci{0.385}{0.358}{0.538} &
\twolineci{0.629}{0.579}{0.766} &
\twolineci{0.267}{0.225}{0.369} &
\twolineci{0.228}{0.190}{0.271} &
\twolineci{0.206}{0.174}{0.290} &
\twolineci{0.420}{0.342}{0.606} \\
{G} & 
\twolineci{0.258}{0.236}{0.371} &
\twolineci{0.525}{0.457}{0.587} &
\twolineci{0.169}{0.132}{0.236} &
\twolineci{0.596}{0.561}{0.624} &
\twolineci{0.152}{0.116}{0.221} &
\twolineci{0.443}{0.281}{0.517} \\
\bottomrule
\end{tabular}
\end{table}

\clearpage

\begin{table*}[htbp]
\centering
\small
\caption{Detailed results of the main evaluation for Task 1 across all categories and participating teams, measured by mean Average Precision (mAP).}
\label{tab:class_performance_TASK1}
\begin{tabular}{lcccccc}
\toprule
\textbf{Category} & \textbf{A} & \textbf{B} & \textbf{C} & \textbf{D} & \textbf{E} & \textbf{F} \\
\midrule
Normal & 0.873 & 0.870 & 0.835 & 0.852 & 0.796 & 0.780 \\
Aortic elongation & 0.626 & 0.634 & 0.616 & 0.572 & 0.561 & 0.467 \\
Cardiomegaly & 0.757 & 0.749 & 0.719 & 0.713 & 0.664 & 0.650 \\
Pleural effusion & 0.820 & 0.823 & 0.751 & 0.785 & 0.758 & 0.586 \\
Nodule & 0.479 & 0.459 & 0.389 & 0.458 & 0.309 & 0.180 \\
Atelectasis & 0.533 & 0.520 & 0.433 & 0.497 & 0.362 & 0.195 \\
Pleural thickening & 0.492 & 0.517 & 0.447 & 0.391 & 0.321 & 0.162 \\
Aortic atheromatosis & 0.356 & 0.358 & 0.211 & 0.219 & 0.226 & 0.178 \\
Support devices & 0.965 & 0.962 & 0.970 & 0.940 & 0.928 & 0.767 \\
Alveolar pattern & 0.590 & 0.589 & 0.526 & 0.585 & 0.527 & 0.454 \\
Fracture & 0.429 & 0.449 & 0.373 & 0.310 & 0.273 & 0.068 \\
Hernia & 0.847 & 0.789 & 0.802 & 0.421 & 0.534 & 0.205 \\
Emphysema & 1.000 & 0.121 & 0.255 & 0.503 & 0.126 & 0.036 \\
Azygos lobe & 0.946 & 0.958 & 0.892 & 0.012 & 0.693 & 0.017 \\
Hydropneumothorax & 1.000 & 0.013 & 0.003 & 0.167 & 0.009 & 0.008 \\
Kyphosis & 0.190 & 0.200 & 0.163 & 0.110 & 0.211 & 0.100 \\
Mass & 0.129 & 0.041 & 0.066 & 0.068 & 0.060 & 0.029 \\
Pneumothorax & 0.608 & 0.143 & 0.437 & 0.689 & 0.076 & 0.013 \\
Subcutaneous emphysema & 0.833 & 0.583 & 0.500 & 0.700 & 0.129 & 0.148 \\
Pneumoperitoneum & 0.571 & 0.146 & 0.200 & 0.013 & 0.518 & 0.006 \\
Vascular hilar enlargement & 0.314 & 0.301 & 0.296 & 0.204 & 0.268 & 0.162 \\
Vertebral degenerative changes & 0.273 & 0.293 & 0.239 & 0.166 & 0.172 & 0.160 \\
Hyperinflated lung & 0.304 & 0.307 & 0.250 & 0.242 & 0.259 & 0.207 \\
Interstitial pattern & 0.538 & 0.580 & 0.478 & 0.591 & 0.469 & 0.346 \\
Central venous catheter & 0.931 & 0.965 & 0.954 & 0.837 & 0.870 & 0.384 \\
Hypoexpansion & 0.368 & 0.457 & 0.330 & 0.267 & 0.304 & 0.153 \\
Bronchiectasis & 0.155 & 0.182 & 0.113 & 0.174 & 0.120 & 0.096 \\
Hemidiaphragm elevation & 0.563 & 0.535 & 0.492 & 0.403 & 0.410 & 0.276 \\
Sternotomy & 0.899 & 0.872 & 0.924 & 0.931 & 0.870 & 0.239 \\
Calcified densities & 0.172 & 0.063 & 0.130 & 0.072 & 0.028 & 0.016 \\
\midrule
\textbf{Mean} & \textbf{0.585} & \textbf{0.483} & \textbf{0.460} & \textbf{0.430} & \textbf{0.395} & \textbf{0.236} \\

\bottomrule
\end{tabular}

\end{table*}

\clearpage

\begin{table}[htbp]
\centering
\caption{Detailed results of the main evaluation for Task 2 across all categories and participating teams, measured by mean Average Precision (mAP).}
\label{tab:task2_results}
\small
\begin{tabular}{lccc}
\toprule
\textbf{Category} & \textbf{A} & \textbf{B} & \textbf{G} \\
\midrule
Scoliosis & 0.878 & 0.737 & 0.606 \\
Osteopenia & 0.127 & 0.069 & 0.079 \\
Bulla & 0.640 & 0.448 & 0.253 \\
Infarction & 0.023 & 0.012 & 0.020 \\
Adenopathy & 0.305 & 0.419 & 0.291 \\
Goiter & 0.617 & 0.178 & 0.091 \\
\midrule
\textbf{Mean} & \textbf{0.432} & \textbf{0.311} & \textbf{0.223} \\
\bottomrule
\end{tabular}

\end{table}

\clearpage

\begin{table}[H]
\centering
\caption{
Descriptive comparison of the Task~2 PadChest-GR internal test set 
and NIH ChestX-ray external test set. Percentages were calculated 
within each evaluation subset.
}
\label{tab:intercenter_heterogeneity}

\footnotesize
\begin{threeparttable}
\begin{tabularx}{\textwidth}{
    X
    l
    l
}
\toprule
\textbf{Characteristic} &
\textbf{PadChest-GR internal test} &
\textbf{NIH ChestX-ray external test} \\
\midrule

\rowcolor{GreenYellow}\multicolumn{3}{l}{\textbf{Label composition}} \\

Number of images
& 305
& 79 \\

Per-class prevalence \\

\hspace{1em}Scoliosis
& 245 (80.3\%)
& 28 (35.4\%) \\

\hspace{1em}Osteopenia
& 13 (4.3\%)
& 1 (1.3\%) \\

\hspace{1em}Bulla
& 15 (4.9\%)
& 21 (26.6\%) \\

\hspace{1em}Infarction
& 0 (0.0\%)
& 3 (3.8\%) \\

\hspace{1em}Adenopathy
& 10 (3.3\%)
& 15 (19.0\%) \\

\hspace{1em}Goiter
& 25 (8.2\%)
& 11 (13.9\%) \\

Images with one positive label
& 302 (99.0\%)
& 79 (100.0\%) \\

Images with two positive labels
& 3 (1.0\%)
& 0 (0.0\%) \\

\midrule
\rowcolor{GreenYellow}\multicolumn{3}{l}{\textbf{Acquisition and metadata}} \\

Projection/view position
& PA: 300 (98.4\%) & PA: 65 (82.3\%)\\
& AP: 5 (1.6\%) & AP: 14 (17.7\%) \\

Modality
& CR: 184 (60.3\%) & --\\
& DX: 121 (39.7\%) & -- \\

Manufacturer
& Philips Medical Systems: 246 (80.7\%) & -- \\
& Imaging Dynamics Company Ltd.: 59 (19.3\%) & -- \\

Original pixel spacing
& --
& $0.147 \pm 0.016 mm \times
0.147 \pm 0.016 mm$ \\

\midrule
\rowcolor{GreenYellow}\multicolumn{3}{l}{\textbf{Image characteristics}} \\

Width, pixels, mean $\pm$ SD
& $2690.9 \pm 522.2$
& $1024.0 \pm 0.0$ \\

Height, pixels, mean $\pm$ SD
& $2534.8 \pm 557.0$
& $1024.0 \pm 0.0$ \\

Aspect ratio, mean $\pm$ SD
& $1.073 \pm 0.101$
& $1.000 \pm 0.000$ \\

Resolution, megapixels, mean $\pm$ SD
& $7.09 \pm 2.87$
& $1.05 \pm 0.00$ \\

Mean grayscale intensity
& $0.397 \pm 0.048$
& $0.450 \pm 0.065$ \\

Grayscale intensity SD
& $0.282 \pm 0.021$
& $0.246 \pm 0.038$ \\

\bottomrule
\end{tabularx}

\begin{tablenotes}[para]
\footnotesize
\item \textit{Note.}
Grayscale intensities were normalized to the range $[0,1]$ before
summary statistics were calculated. Formal image-quality ratings were
not available for either dataset; resolution and intensity statistics
are therefore reported as directly measurable image characteristics
rather than formal measures of diagnostic image quality.
A dash indicates that the corresponding metadata were unavailable.
PA: posteroanterior; AP: anteroposterior; CR: computed radiography;
DX: digital radiography; SD: standard deviation.
\end{tablenotes}
\end{threeparttable}
\end{table}

\end{document}